\documentclass[10pt,journal,compsoc]{IEEEtran}
\usepackage{caption}
\usepackage{wrapfig}
\usepackage{graphicx}
\usepackage{amsmath}
\usepackage{amssymb}
\usepackage{booktabs}
\usepackage{makecell}
\usepackage{multirow}
\usepackage{colortbl}
\usepackage{bbding}

\ifCLASSOPTIONcompsoc
  \usepackage[nocompress]{cite}
\else
  \usepackage{cite}
\fi

\ifCLASSINFOpdf
\else
\fi
\usepackage{amsmath}

\usepackage{algorithm}

\ifCLASSOPTIONcompsoc
 \usepackage[caption=false,font=footnotesize,labelfont=sf,textfont=sf]{subfig}
\else
 \usepackage[caption=false,font=footnotesize]{subfig}
\fi

\usepackage{url}

\usepackage{listings}

\begin{document}
\def\etal{\emph{et al}.}
\def\eg{\emph{e}.\emph{g}.}
\def\ie{\emph{i}.\emph{e}.}
\definecolor{Gray}{gray}{0.8}
\newcolumntype{g}{>{\columncolor{Gray}}c}

\title{
A Unified Multimodal \textit{De}- and \textit{Re}-coupling Framework for RGB-D Motion Recognition 
}
%
%
%
%

\author{Benjia~Zhou,
        Pichao Wang,
        Jun Wan,
        Yanyan Liang, %
        Fan Wang
\thanks{Manuscript received April 26, 2023; revised March 03, 2023. Corresponding authors: Pichao Wang and Jun Wan.}
\IEEEcompsocitemizethanks{\IEEEcompsocthanksitem  B. Zhou and Y. Liang are with Macau University of Science and Technology, Macau 999078, China (e-mail: 21098536ia30001@student.must.edu.mo;
yyliang@must.edu.mo).
\IEEEcompsocthanksitem  P. Wang and F. Wang are with DAMO Academy, Alibaba Group (U.S.) Inc., Bellevue, WA, 98004, USA (e-mail: pichaowang@gmail.com, fan.w@alibaba-inc.com). Pichao Wang is now affiliated with Amazon.
\IEEEcompsocthanksitem J. Wan is with the State Key Laboratory of Multimodal Artificial Intelligence Systems (MAIS), Institute of Automation, Chinese Academy of Sciences (CASIA), Beijing 100190, China, also with the School of Artificial Intelligence, University of Chinese Academy of Sciences (UCAS), Beijing 100049, China, and also with Macau University of Science and Technology, Macau 999078, China (e-mail: jun.wan@ia.ac.cn).

}
}

\markboth{IEEE Transactions on Pattern Analysis and Machine Intelligence}%
{Shell \MakeLowercase{\textit{et al.}}: Bare Demo of IEEEtran.cls for Computer Society Journals}


\IEEEtitleabstractindextext{%
\begin{abstract}
Motion recognition is a promising direction in computer vision, but the training of video classification models is much harder than images due to insufficient data and considerable parameters. To get around this, some works strive to explore multimodal cues from RGB-D data. Although improving motion recognition to some extent, these methods still face sub-optimal situations in the following aspects: 
(i) Data augmentation, \ie, the scale of the RGB-D datasets is still limited, and few efforts have been made to explore novel data augmentation strategies for videos;
(ii) Optimization mechanism, \ie, the tightly space-time-entangled network structure brings more challenges to spatiotemporal information modeling; 
And (iii) cross-modal knowledge fusion, \ie, the high similarity between multimodal representations leads to insufficient late fusion.
To alleviate these drawbacks, we propose to improve RGB-D-based motion recognition both from data and algorithm perspectives in this paper. 
In more detail, firstly, we introduce a novel video data augmentation method dubbed ShuffleMix, which acts as a supplement to MixUp, to provide additional temporal regularization for motion recognition. 
Secondly, a \textbf{U}nified \textbf{M}ultimodal \textbf{D}e-coupling and multi-stage \textbf{R}e-coupling framework, termed UMDR, is proposed for video representation learning.
Finally, a novel cross-modal Complement Feature Catcher (CFCer) is explored to mine potential commonalities features in multimodal information as the auxiliary fusion stream, to improve the late fusion results.
The seamless combination of these novel designs forms a robust spatiotemporal representation and achieves better performance than state-of-the-art methods on four public motion datasets. Specifically, UMDR achieves unprecedented improvements of $\uparrow4.5\%$ on the Chalearn IsoGD dataset. Our code can be available at \url{https://github.com/zhoubenjia/MotionRGBD-PAMI}.
\end{abstract}

}

\maketitle

\IEEEdisplaynontitleabstractindextext

%
\IEEEpeerreviewmaketitle

\IEEEraisesectionheading{\section{Introduction}\label{sec:introduction}}
\IEEEPARstart{H}{uman} motion recognition~\cite{Carreira_2017_CVPR,Abavisani_2019_CVPR,duan2018unified} 
is an attractive yet challenging task. Recently, RGB-D-based motion recognition~\cite{wan2013one,Wang_2017_ICCV,2018Depth} has attracted much attention in computer vision due to the easy availability of depth data and its extensive application prospects such as video surveillance and human-object interfaces. RGB-D-based motion recognition refers to jointly exploring the color and depth cues in multimodal data by constructing a multi-stream processing mechanism, thereby forming an informative joint representation for motion description~\cite{wan2015explore, wang2018cooperative, de2020infrared}. 

\begin{figure}[!t]
\centering
\subfloat[NTU-RGBD]{\includegraphics[width=0.5\linewidth]{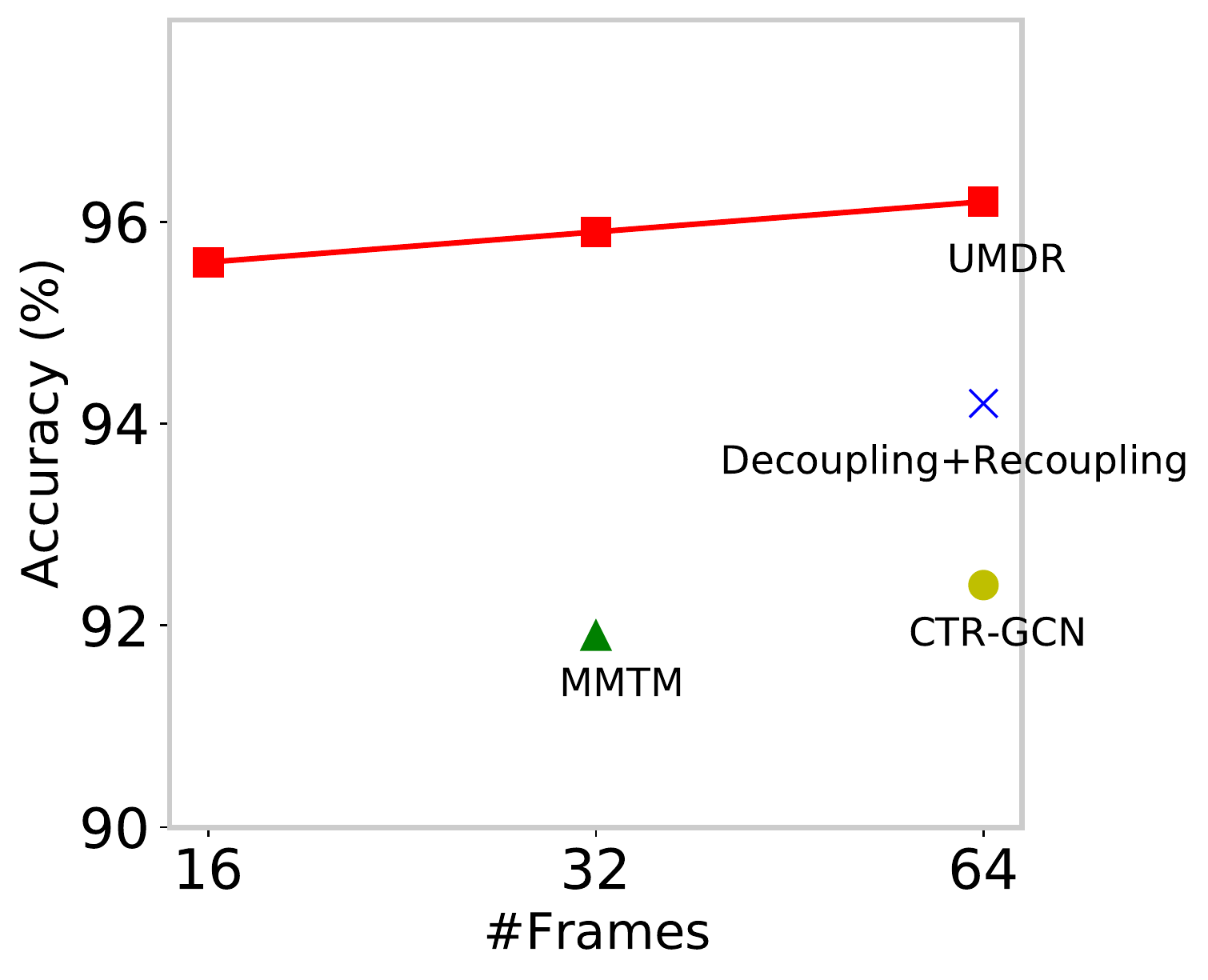}}
\hfil
\subfloat[IsoGD]{\includegraphics[width=0.5\linewidth]{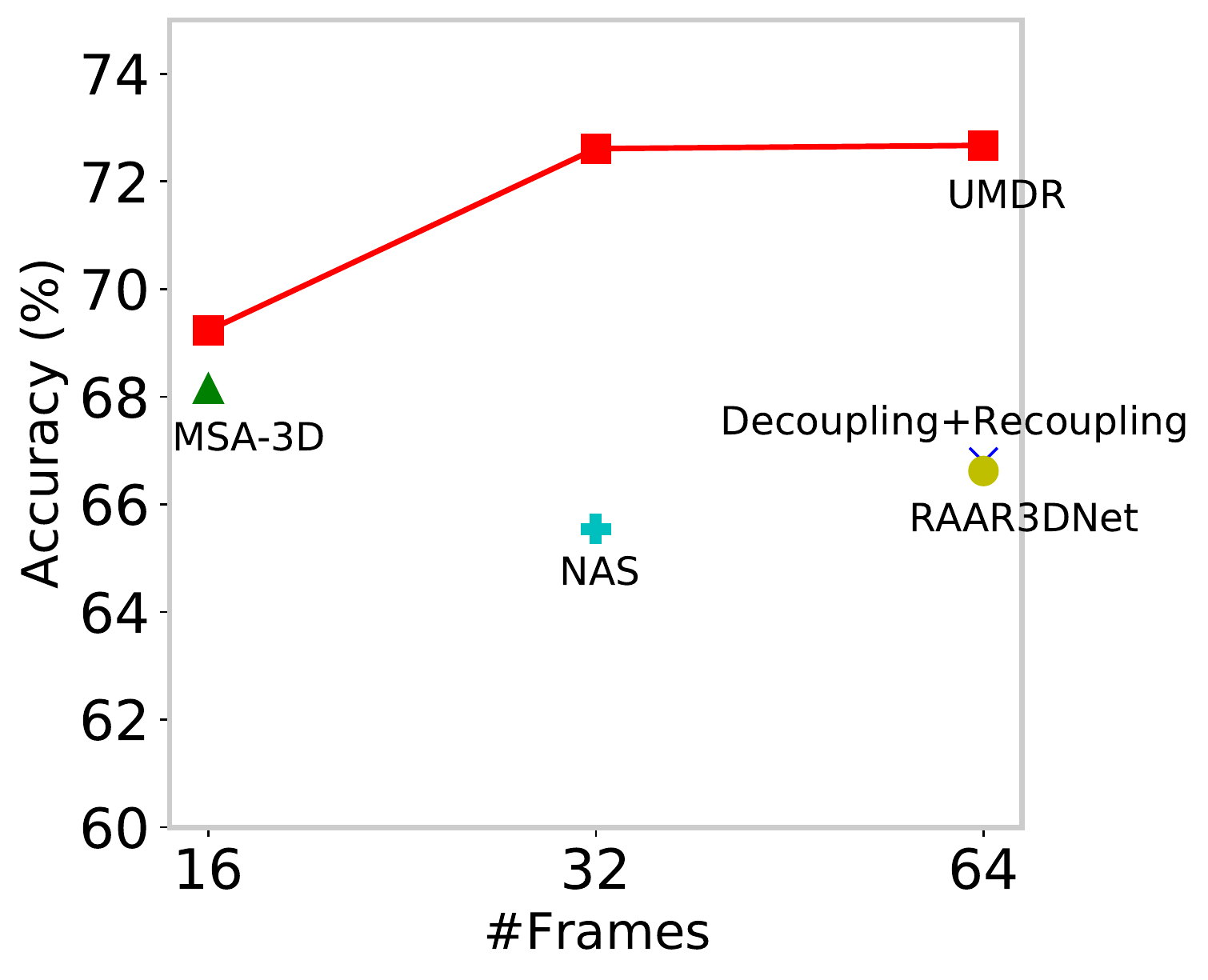}}
\caption{\textbf{UMDR outperforms a number of state-of-the-art methods on both action and gesture datasets.}}
\label{fig:sota-compa}
\end{figure}
\begin{figure}[!t]
    \centering
    \includegraphics[width=1.0\linewidth]{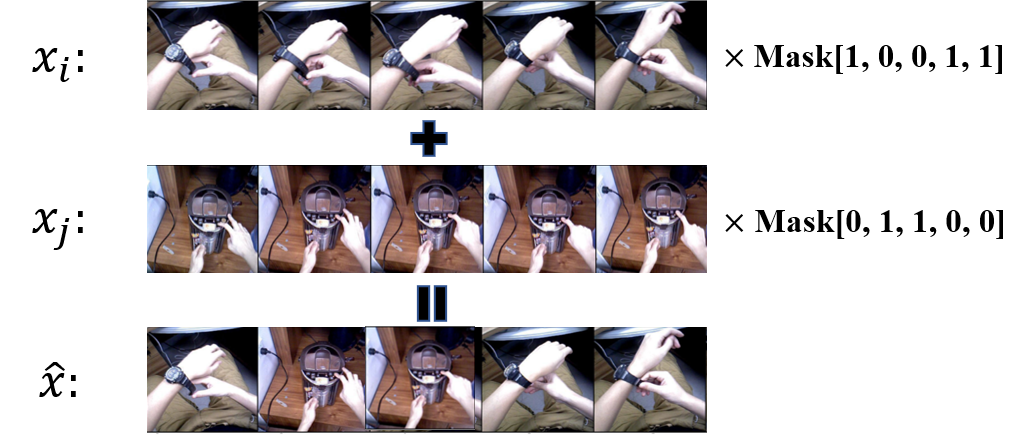}
    \caption{\textbf{The proposed video ShuffleMix strategy.} Unlike MixUp, we aim to mix two videos in the time dimension, thereby enhancing the temporal perception.}
    \label{fig:ShuffleMix}
\end{figure}

At present, although various architectures ~\cite{Joze_2020_CVPR, yu2021searching, li2021trear, Zhou_Li_Wan_2021} have been continuously suggested to improve the multi-modal motion recognition (as shown in Fig.\ref{fig:sota-compa}), the majority of cutting-edge works strive to improve the performance at the algorithm level, ignoring the importance of data-level improvements. 
In addition, MixUp~\cite{zhang2018mixup} shines in image classification tasks as a simple yet effective data augmentation method. However, directly transferring the MixUp strategy to the video classification task falls into sub-optimal due to the spatiotemporal-entangled nature of video data.
Therefore, in this paper, we gain insight into the role of MixUp in the video domain first, and then propose an alternative counterpart of it, namely ShuffleMix, as illustrated in Fig.\ref{fig:ShuffleMix}. Concretely, it shuffles two video clips together in the temporal dimension to form a new instance, while maintaining their respective temporal consistency. The experimental results show that the best performance gain can be obtained by combining it with the MixUp strategy, termed ShuffleMix+ in this paper.

\begin{figure*}[!t]
\centering
\subfloat[Multimodal Training Framework]{\includegraphics[width=0.6\linewidth]{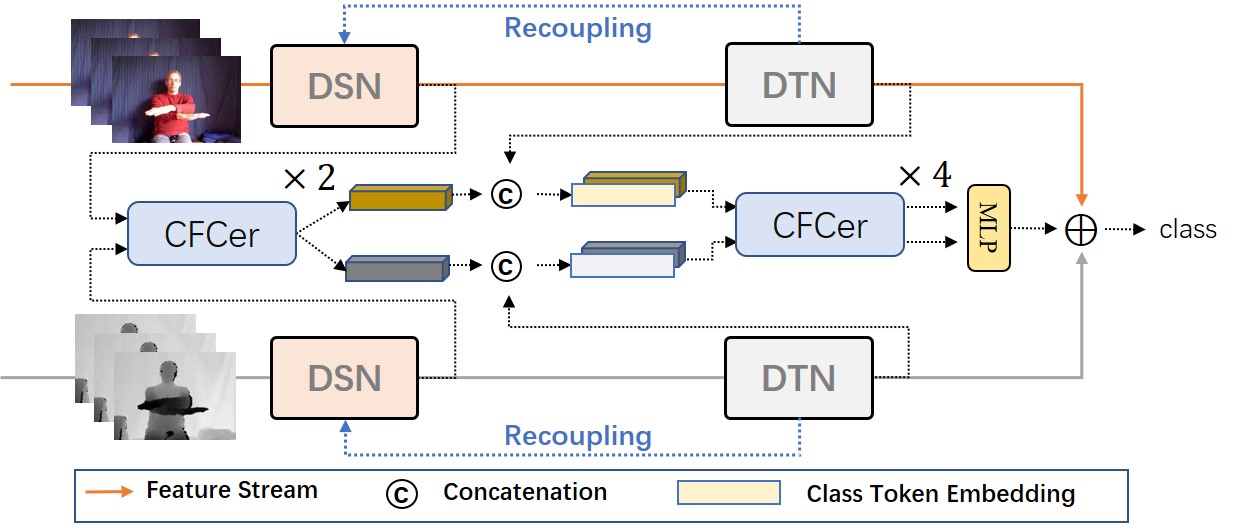}%
\label{fig:CAPFM}}
\hfill
\subfloat[Complement Feature Catcher (CFCer)]{\includegraphics[width=0.4\linewidth]{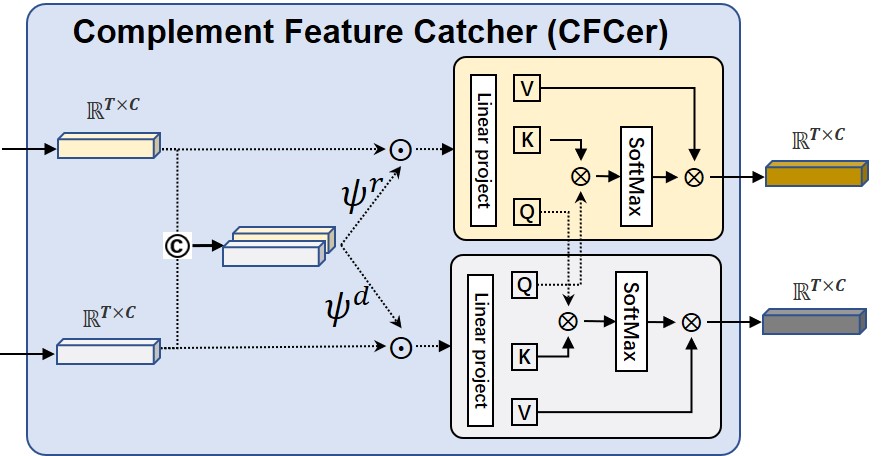}%
\label{subfig:cfc}}
\caption{\textbf{Overview of proposed method.} (a) The multimodal \textit{De-}/\textit{Re-}recoupling learning pipeline and fusion mechanism. For the unimodal representation learning, we learn the dimension-specific (spatial/temporal) features by the well-designed spatial and temporal networks DSN and DTN; For multimodal knowledge fusion, we modify the late fusion by separating the modality-specific complementary features along spatial to temporal domains.
(b) The structure of the proposed CFCer. It contains feature enhancement and separation mechanisms that can mine the distinguishing complementary correspondences from multimodal signals.
}
\label{fig:framework}
\end{figure*}

In addition, we notice that most of the recent works capture spatiotemporal correspondences through the space-time coupled modeling structure (\eg, based on a 3D convolution kernel \cite{Carreira_2017_CVPR, yu2021searching, Zhou_Li_Wan_2021}), inevitably introducing the following challenges: 
(i) Optimization difficulty arises in the case of limited RGB-D data due to the tightly coupled nature of the spatiotemporal dimension, implying that spatiotemporal information affects each other when performing features aggregation;
(ii) Spatiotemporal redundancy information is hard to filter in the entangled space-time space, leading to decreased efficiency and robustness of the model; 
Although some works~\cite{tran2018closer, kalfaoglu2020late,miao2021spatiotemporal} attempt to decompose the 3D modeling procedure to decouple spatial and temporal learning, they are still problematic in terms of compact representation learning, \ie, they somewhat weaken or even destroy the original spatiotemporal coupling structure, resulting in losing the spatiotemporal dependency in a video. In our case, as illustrated in Fig.\ref{fig:framework}, we propose a Decoupled Spatial representation learning Network (DSN) and Decoupled Temporal representation learning Network (DTN) to learn the dimension-specific representations, which can reduce the optimization difficulty of modeling 3D information since DSN and DTN capture spatial and temporal features in their respective roles. Additionally, we introduce a multi-stage recoupling mechanism to reinforce the spatiotemporal connection by re-establishing the time-space correspondence.

Furthermore, late fusion is still popular in RGB-D-based motion recognition \cite{zhu2019redundancy, Abavisani_2019_CVPR, yu2021searching,  Zhou_Li_Wan_2021} due to its flexibility and effectiveness. However, we observe a high similarity between multimodal spatiotemporal representations learned from individual color and depth signals, which could result in suboptimal fusion outcomes due to the fact that it forces two independent classifiers to converge to the same decision surface.
Therefore, we design a novel late fusion mechanism as illustrated in Fig.\ref{fig:framework}, which improves the fusion results by capturing fine-grained complementary correspondences from multimodal signals as an auxiliary fusion stream via the proposed Complement Feature Catcher (CFCer). This is due to the fact that we observe that the cross-modal semantic complementary features retain a high distinctiveness from the unimodal spatiotemporal representation.

In more detail, we improve motion recognition from the following three aspects:
(1) \textbf{Learning with spatiotemporal regularization.}
First, the video clips are preprocessed by ShuffleMix+, which randomly mixes example pairs in both spatial and temporal dimensions, greatly increasing the diversity of inputs. Meanwhile, the corresponding label pairs are also mixed together as the supervision signal. Subsequently, the generated mixture samples and labels are used for training. 
(2) \textbf{Learning with decoupling followed by recoupling.}
The mixture samples are first input into the DSN for hierarchical spatial features extraction and reinforcement by a stack of multi-scale features learning modules (SMS) and recoupling modules (RCM), respectively. Then, we sample several sub-sequences at different frame rates from the enhanced spatial features through a Sampling Layer (SL) as the input of DTN. The DTN captures the hierarchical local fine-grained and global coarse-grained temporal features through a multi-branch structure with a temporal multi-scale features learning module (TMS) and Multi-branch Transformer blocks (MTrans). Additionally, a multi-stage recoupling strategy, as depicted by the blue dashed line in Fig.\ref{fig:pipline}, is developed to reinforce the space-time interdependence.
(3) \textbf{Fusion with cross-modal complementary features.}
The multimodal feature streams are fed into a stack of CFCers to separate the modality-specific spatiotemporal complementary correspondences along the spatial to temporal domains, thereby aiding in classification.

This work is an extension of our CVPR 2022 paper, and the extensions over the original conference version~\cite{Zhou_2022_CVPR} include: 
(1) We introduce a new video data augmentation strategy for RGB-D motion recognition.
(2) We extend the RCM module as a multi-head mechanism, which enables it to generate a discriminative attention map for each video frame. And we extend the single-stage recoupling strategy to a multi-stage strategy.
(3) We gain insight into the factors that influence multimodal late fusion, and improve fusion results by introducing modality-specific complementary features.
(4) We achieve state-of-the-art results on four commonly used RGB-D motion datasets.

\section{Related Work}
\subsection{Video Data Augmentation}
Data augmentation, a data-driven and informed regularization strategy that artificially increases the number of training samples to improve the generalization ability of models, has been widely explored for better training of deep neural networks~\cite{shorten2019survey,touvron2021training,balestriero2022effects,xu2022comprehensive}.
Generally speaking, video data augmentation can be roughly divided into two categories, namely, single-sample-based and multi-sample-based. 
The former refers to conducting data augmentation around the sample itself, including geometric transformation, color transformation, and frame sampling.
Joe \etal~\cite{yue2015beyond} leverage an upsampling strategy to significantly improve the performance on human action recognition tasks. Hoai \etal~\cite{hoai2014improving} use a novel data augmentation by mirroring the videos. Basura \etal~\cite{Fernando_2015_CVPR} attempt to learn video-wide representation by arranging frames in chronological order. Furthermore, \cite{ok2021vidaug} provides a practical single-sample Video Augmentation library for Deep Learning architectures. 
The latter refers to utilizing multiple samples to generate a new instance, including video MixUp~\cite{zhang2018mixup} and CutMix~\cite{yun2019cutmix}.
Yun \etal~\cite{yun2020videomix} propose a new video augmentation strategy, VideoMix, which generates a new instance by inserting a video cube into another one. Wang \etal~\cite{wang2021removing} selects a static frame and adds it to every other frame to construct a distracting video, thereby alleviating the problem of \textit{implicit biases} that commonly exists in video datasets. Wu \etal~\cite{wu2022dynamic} mix two videos from different domains with fixed weights, to tackle the cross-domain problem in video.
Our approach is based on the second data augmentation strategy, where we try to mix two video clips together in both the space and time dimensions to reinforce the spatiotemporal perception.

\subsection{Motion Recognition based on RGB-D Data}
\label{sec:rgbd}
Human motion recognition is more challenging than image classification due to the optimization difficulty and data scarcity. RGB-D-based motion recognition~\cite{zhu2016large,Chai2016two, wang2017scene, Zhu2017multimodal, wang2018cooperative,Shahroudy2018deep,li2019large, ZHOU2021235, Zhou_Li_Wan_2021,Zhou_2022_CVPR}, thus, has attracted extensive attention thanks to the availability of additional depth cues. For example, 
Zhu \etal~\cite{zhu2016large} leverage pyramid-like input and fusion layers to preserve and fuse multi-scale contextual features.
Ye \etal~\cite{Ye_2016_CVPR_Workshops} propose a Spatiotemporal-LSTM framework to learn motion patterns from temporal information with dense motion trajectories.
Kong \etal~\cite{kong2015bilinear} attempt to project the RGB-D data into a shared space to learn cross-modal features. 
Yu \etal~\cite{yu2021searching} construct a cross-modal adaptive multi-branch network to learn and fuse multimodal representations via NAS. Scene flow is adopted in \cite{wang2017scene} for compact RGB-D representation learning.
Different from others, Wang \etal~\cite{wang2018cooperative} attempt to use a single convolution neural network (dubbed c-ConvNet) to learn multimodal spatiotemporal representation.
In addition, Zhang \etal~\cite{zhang2020hierarchically} present a hierarchically decoupled spatiotemporal contrastive learning method for self-supervised video representation learning. He \etal~\cite{he2019stnet} propose an effective spatiotemporal network StNet, which employs separated channel-wise and temporal-wise convolution operations for decoupled local and global representation learning.
In this paper, we aim to learn recoupled features to exploit the strong correlations between time and space, and to improve multimodal fusion results by learning cross-modal complementary knowledge in RGB-D data along the spatial to temporal domains, which is important but has actually received little attention.

\begin{figure*}[t]
\centering
\includegraphics[width=1.0\linewidth]{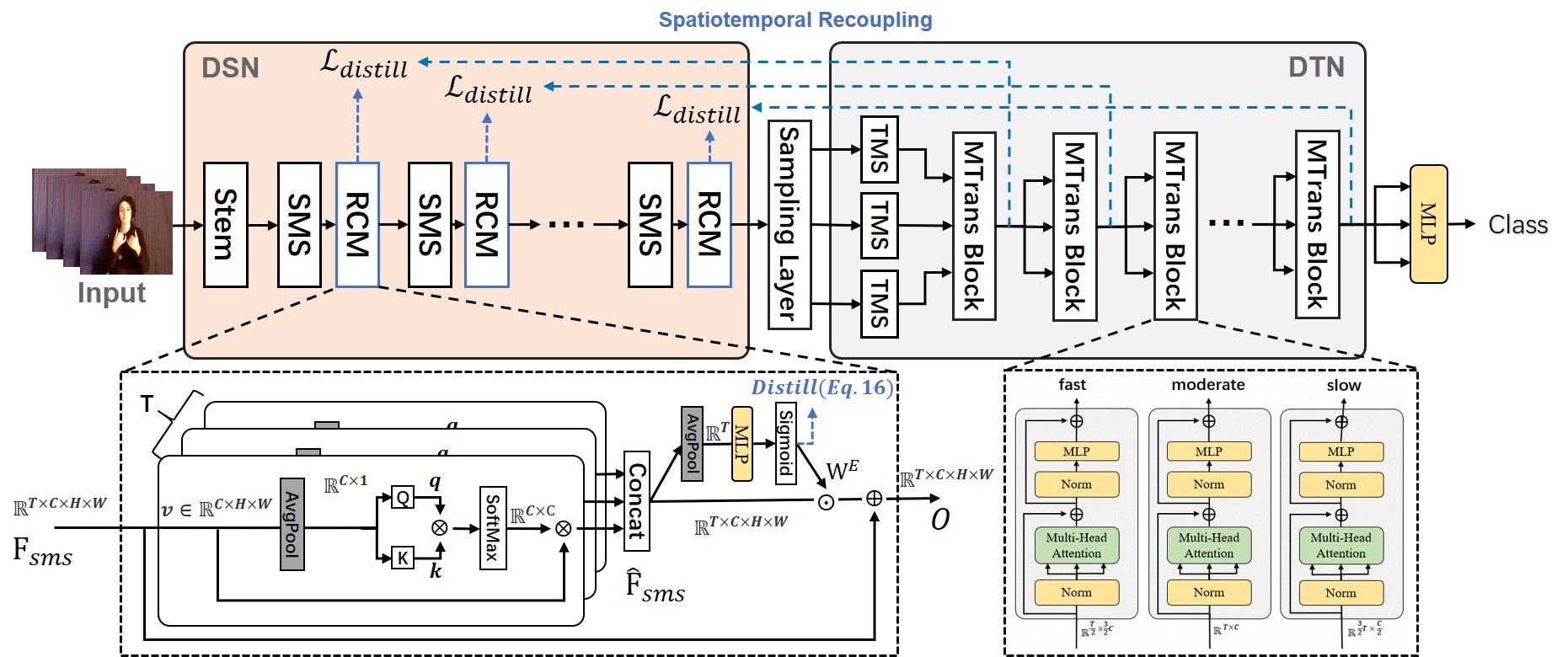}
\caption{\textbf{The unimodal representation learning network.} 
In DSN, the \textit{Steam} layer first extracts low-level visual embeddings from the input. They are then fed into the SMS module to construct hierarchical spatial features, and the RCM module for feature selection by calculating dynamic channel attention maps and distilling temporal knowledge. In DTN, the spatial features are fed into the TMS module and MTrans Blocks to learn local fine-grained and global coarse-grained temporal features.
}
\label{fig:pipline}
\end{figure*}
\section{Proposed Method}
In this paper, we improve the RGB-D motion recognition both from the data and algorithm perspectives. For the former, we introduce a novel video augmentation method to enlarge the training data space (Sec.\ref{sec:ShuffleMix}). For the latter, we first decompose the spatiotemporal modeling procedure for dimension-independent learning. Then introduce a multi-stage recoupling method to reestablish the time-space association (Sec.\ref{sec:drcoup}). Finally, we improve the multimodal late fusion by exploring the modality-specific complementary features (Sec.\ref{sec:fusion}). 

\subsection{Spatiotemporal Regularization}
\label{sec:ShuffleMix}
In this section, we first investigate the existing data augmentation MixUp~\cite{zhang2018mixup} in video classification, and then introduce the proposed ShuffleMix and its variant ShuffleMix+. 

\noindent\textbf{Video MixUp.}\quad
Given two video clips $V_i=\{v_{i,1}, v_{i,2}, \cdots, v_{i,T}\}$ and $V_j=\{v_{j,1}, v_{j,2}, \cdots, v_{j,T}\}$ of length $T$. In a nutshell, MixUp constructs virtual input/output pairs ($\hat{V}$, $\hat{y}$) for training:
\begin{equation}
    \left\{
\begin{aligned}
&\hat{V}  =  \lambda_m \cdot V_i+(1-\lambda_m)\cdot V_j, \quad \hat{V}\in \mathbb{R}^{T\times 3 \times H \times W},\\
&\hat{y}  =  \lambda_m \cdot y_i + (1-\lambda_m) \cdot y_j, \quad \hat{y}\in \mathbb{R}^C.
\end{aligned}
\right.
\end{equation} 
where $\lambda_m \sim Beta_{[0, 1]}(\alpha_m, \alpha_m)$, $\alpha_m$ is a hyper-parameter that controls the strength of interpolation between sample pairs,  
$y_i$ and $y_j$ denote one-hot label encodings, and $C$ is the total number of classes.
In image classification, according to \cite{carratino2020mixup} and \cite{zhang2021how}, training with MixUp can be approximated as optimization with a spatial regularization term to improve the generalization and robustness of models. 
Taking the cross-entropy loss $l(\cdot, \cdot)$ as an example, the loss function based on MixUp can be described as (refer to ~\cite{carratino2020mixup} for more):
\begin{equation}
    \label{eq:mixup_loss}
    \begin{split}
         \xi^{mixup}(f) = &\frac{1}{N^2}\sum_{i=1}^N{\sum_{j=1}^N{\mathbb{E}_{\lambda_m} ~ l(\hat{y}, f(\hat{V}))}} \\
         =&\frac{1}{N}\sum^{N}_{i=1}{\mathbb{E}_{\theta, j}~l(\tilde{y}_i+\epsilon_i, f(\tilde{V}_i+\delta_i))}
    \end{split}
\end{equation}
where $\theta \sim Beta_{[\frac{1}{2}, 1]}(\alpha, \alpha)$,  $j\sim Unif[N]$, $f(\cdot)$ is feature extractor, $N$ is batch size. ($\tilde{V}_i$, $\tilde{y}_i$) means the input/output pair mapped from ($\hat{V}$, $\hat{y}$) through $\Bar{\theta}\in [1/2, 1]$, which are given by:
\begin{equation} \label{eq:mapping}
    \left\{
\begin{aligned}
&\tilde{y}_i  =  \Bar{y}+\Bar{\theta}(y_i-\Bar y), \\
&\tilde{V}_i  =  \Bar{V}+\Bar{\theta}(V_i-\Bar{V}),
\end{aligned}
\right.
\end{equation}
where $\Bar{\theta}=\mathbb{E}_\theta \theta$, $\Bar{V}=\frac{1}{N}\sum_{i=1}^N{V_i}$, and $\Bar{y}=\frac{1}{N}\sum_{i=1}^N{y_i}$.
Furthermore, $\delta_i$ and $\epsilon_i$ represent random spatial perturbations of input and output, respectively. They are given by:
\begin{equation} \label{eq:pertur}
    \left\{
\begin{aligned}
&\delta_i  =  (\theta - \Bar{\theta})V_i+(1-\theta)V_j-(1-\Bar{\theta})\Bar{V}, \\
&\epsilon_i  =  (\theta - \Bar{\theta})y_i+(1-\theta)y_j-(1-\Bar{\theta})\Bar{y},
\end{aligned}
\right.
\end{equation}
where $\mathbb{E}_{\theta, j}\delta_i=\mathbb{E}_{\theta, j}\epsilon_i=0$.
From the above observations, we find standard MixUp treats the entire video sequence as a still image with $T \times 3$ channels and adds spatial noise perturbations over it. 
However, we argue that this is a suboptimal solution for MixUp's expansion into the video domain because no additional temporal noise perturbation is injected. To fill this deficiency of MixUp, we introduce a new temporal regularization method, termed ShuffleMix, which acts as a supplement to MixUp, to help achieve regularization at the spatiotemporal level.

\begin{figure}[!t]
\centering
\subfloat[Local-based Continuous Mixture.]{\includegraphics[width=3in]{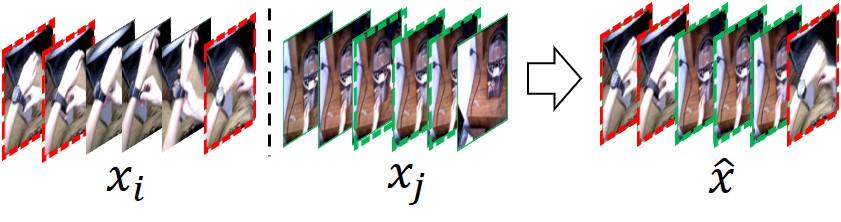}%
\label{subfig:mix_local}}
\hfil
\subfloat[Global-based Discrete Mixture.]{\includegraphics[width=3in]{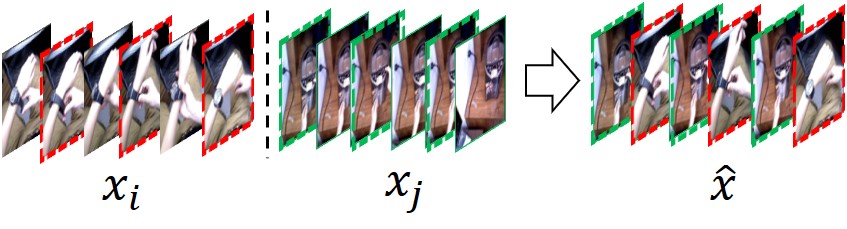}%
\label{subfig:mix_global}}
\caption{\textbf{Two variants of video mixture strategies.} The (a) preserves local temporal consistency, whereby encouraging the exploration of fine-grained temporal cues; The (b) ignores temporal consistency in video clips, whereby relaxing the strong temporal dependence in motion recognition.}
\label{fig:mix_type}
\end{figure}
\noindent\textbf{Video ShuffleMix.}\quad
ShuffleMix learns the motion patterns from the randomly shuffled video pairs, which captures how specific acts are related to a temporal context. More specifically, it mixes two video clips $V_i$ and $V_j$ along the temporal dimension to construct a new instance  while preserving the respective temporal consistency, as shown in Figure.~\ref{fig:ShuffleMix}. 
Formally, the video clips $V_i$ and $V_j$ are first masked along the temporal dimension using the randomly initialized mask matrix $\mathbf{Mask}_{\lambda_s, i}\in\mathbb{R}^{T}$ with a mask rate of $\lambda_s$, and its contrariety matrix $\mathbf{Mask}_{1-\lambda_s, j}\in\mathbb{R}^{T}$ with a mask rate of $1-\lambda_s$ respectively, written as:
\begin{equation} \label{eq:masked}
    \left \{
\begin{aligned}
     & V'_i  =  \mathbf{Mask}_{\lambda_s, i} \cdot V_i, \quad V'_i\in \mathbb{R}^{T \times 3 \times H \times W} \\
     & V'_j  =  \mathbf{Mask}_{1-\lambda_s, j} \cdot V_j,\quad V'_j \in \mathbb{R}^{T \times 3 \times H \times W}
\end{aligned}
\right.
\end{equation}
where $i\neq j$, $\mathbf{Mask}_{1-\lambda_s, j} = 1-\mathbf{Mask}_{\lambda_s, i}$. 
Then taking the $V'_i$ and $V'_j$ as a set of new input pairs for training, written as:
\begin{equation} \label{eq:shuffmix}
    \left \{
\begin{aligned}
     & \check{V}  = V'_i + V'_j, \quad \check{V} \in \mathbb{R}^{T \times 3 \times H \times W} \\
     & \check{y}  =  \lambda_s \cdot y_i + (1-\lambda_s) \cdot y_j, \quad \check{y} \in \mathbb{R}^{C}
\end{aligned}
\right.
\end{equation}
where $\lambda_s \sim Beta_{[0, 1]}(\alpha_s, \alpha_s)$ controls the strength of temporal regularization between sample pairs. 
After that, 
the loss function based on ShuffleMix can be re-written as:
\begin{equation}
    \label{eq:ShuffleMix_loss}
    \begin{split}
         \xi^{ShuffleMix}(f) = \frac{1}{N^2}\sum_{i=1}^N{\sum_{j=1}^N{\mathbb{E}_{\lambda_s} ~ l(\check{y}, f(\check{V}))}} 
    \end{split}
\end{equation}
According to our research, we find that ShuffleMix is more friendly to compound motions, such as compound gestures that contain multiple complex movement trajectories, while MixUp prefers simple single-trajectory motions. This is because the ShuffleMix enhances fine-grained motion perception by randomly injecting temporal perturbation in the input, which is insignificant in vanilla MixUp. 
Therefore, as outlined in Algorithm.1, to achieve both spatial and temporal regularization, we introduce a hyper-parameter $\rho$ to control MixUp and ShuffleMix to work together, and it degenerates into a standard MixUp when $\rho=0$. We formally name this new regularization method as ShuffleMix+ throughout the paper and thoroughly investigate its impact on performance in Table.\ref{tab:mix_ratio} below.

\begin{algorithm}[t] \label{alo:ShuffleMix+}
\small
\caption{Codes of ShuffleMix+ in Python-like.}
\begin{lstlisting}[language=Python]
class ShuffleMix+(object):
  def __init__(self, alpha_m, alpha_s, rho):
    self.rho = rho
    self.ShuffleMix = ShuffleMix(alpha_s)
    self.mixup = MixUp(alpha_m)
  def __call__(x, target):
    if self.rho > random.uniform(0,1):
        x, target = self.ShuffleMix(x, target)
    else:
        x, target = self.mixup(x, target)
    return x, target
\end{lstlisting}
\end{algorithm}
\noindent\textbf{Video ShuffleMix+.}\quad In practical implementation, we investigate two straightforward strategies to realize ShuffleMix+ during training: (1) each mini-batch data has a $\rho\%$ probability of using ShuffleMix, otherwise MixUp; (2) each example pair has a $\rho\%$ probability of using ShuffleMix, otherwise MixUp. 
Moreover, for temporal mixture, we investigate two types of interpolation methods as depicted in in Fig.\ref{fig:mix_type}. The first one is the normal interpolate approach similar to \cite{yun2020videomix}, which randomly replaces locally consecutive frames in the video clip as shown in Fig.\ref{subfig:mix_local}. The second one is a discrete interpolation approach based on the global perspective, which randomly replaces some frames across the entire sequence without considering the sequence continuity as depicted in Fig.\ref{subfig:mix_global}. 
Among them, the former encourages exploring the fine-grained temporal cues in sequences, while the latter relaxes strong context temporal dependencies in motion recognition. This consideration is based on the following fact: (i) some gestures are represented by subtle movements (\eg, slight wiggles of the fingertips); and (ii) some actions can be directly recognized by some static images in the video without relying heavily on contextual temporal evolution (\eg, drinking water). 
We discuss the performance gains brought by these two temporal mixture strategies in detail in Section~\ref{sec:diffshuff}.

\subsection{Decoupling and Recoupling Learning} \label{sec:drcoup}
Similar to video data augmentation, optimizing the modeling mechanism for video is also a promising research direction. In this paper, we hypothesize that the superior spatiotemporal representation is acquired from spatiotemporal decoupling learning, followed by spatiotemporal recoupling. Over on this assumption, we learn domain-independent features from the decoupled spatial (Sec.\ref{sec:decoup_spai}) and temporal (Sec.\ref{sec:decoup_tempo}) domains, and recouple (Sec.\ref{sec:recoup}) them with self-distillation to form informative spatiotemporal representations for motion recognition.

\subsubsection{Decoupled Spatial Feature Learning (DSN)}\label{sec:decoup_spai}
As shown in Fig.\ref{fig:pipline}, the DSN involves three important components: (i) a \textit{Steam} layer for mapping visual embedding; (ii) a stack of SMS modules for capturing spatial multi-scale features; and (iii) the RCM modules for feature selection as well as time-space recoupling.
Let $V\in \mathbb{R}^{T \times 3 \times H \times W}$ denotes the input with length $T$ sampled from the video. It is first fed into the \textit{Steam} to capture the low-level  texture features.

\noindent\textbf{Steam.}\quad Similar to \cite{Szegedy2017Inceptionv4}, the \textit{Steam} module is the initial set of operations used to map low-level visual embedding before executing to the SMS module, written as:
\begin{equation}
    X = \mathrm{Steam}(V, \Theta), \quad X \in \mathbb{R}^{T \times C \times H \times W},
\end{equation}
where $\Theta$ represents the learnable parameter matrices. Here we employ the first five layers of the I3D~\cite{Carreira_2017_CVPR} as the \textit{Steam} module.
The captured features are then fed into the subsequent SMS modules for hierarchical spatial feature learning.

\noindent\textbf{SMS Module.}\quad The SMS module is composed of a space-centric 3D Inception Module\footnote{The size of the convolution kernel in the temporal dimension is 1.} \cite{Carreira_2017_CVPR} and a Max Pooling layer, which extracts the multi-scale spatial features by:
\begin{equation}
F_{sms} = 
\textrm{Maxpool}(\mathrm{InC}_{1\times k\times k}(X, \Theta))
\end{equation}
where $\mathrm{InC}_{1\times k\times k}(\cdot, \Theta)$ indicates the Inception Module with a kernel size of $1\times k\times k$. Herein we set $k$=3.
The consideration of capturing hierarchical spatial features is motivated by the fact that there are multiple scale variations in space when an action occurs. However, this may incur more spatial redundancy. To tackle this problem, we cascade the recoupling module (RCM) after each SMS to enhance the most significant aspects while suppressing those that are not.

\noindent\textbf{RCM Module.}\quad The RCM module is designed to enhance the spatial features from both the intra-frame (channel) and inter-frame (sequence), as illustrated in Fig.\ref{fig:pipline}. 
To achieve the goal in the former aspect, inspired by \textit{self-attention} \cite{dosovitskiy2020image}, we calculate dynamic attention maps based on the input to re-weight spatial features of each frame to highlight sensitive filters in it. 
Concretely, taking the first RCM module as an example, we first define a set of learnable matrices $\mathbf{W}^Q \in \mathbb{R}^{C\times d \times T}$ and $\mathbf W^K\in \mathbb{R}^{C\times d \times T}$ with $T$ heads of dimension $\mathbb{R}^d$. The RCM takes the $F_{sms}\in \mathbb{R}^{T \times C \times H \times W}$ as the input and dynamically calculate the attention maps $\mathbf A \in \mathbb{R}^{C\times C \times T}$, followed by a softmax activation function normalizing the weight values. Then it is used to re-weight spatial features $F_{sms}$ along the channel dimension. The entire process can be formulated as:
\begin{equation}
    \begin{split}
    & \mathbf q=F_{sms} \cdot \mathbf W^Q,\quad \mathbf k= F_{sms} \cdot \mathbf W^K, \\
    & \mathbf{A} = \textrm{softmax}(\frac{\mathbf q \cdot \mathbf k^T}{\sqrt{d}}), \\
    & \hat{F}_{sms} = \mathbf{A} \cdot F_{sms}, \quad  \hat{F}_{sms} \in \mathbb{R}^{T \times C \times H \times W}.
    \end{split}
\end{equation}
where $\hat{F}_{sms}$ is the intra-frame enhanced spatial features, $\mathbf q$ and $\mathbf k$ denote the queries and keys, respectively. Unlike the standard \textit{self-attention}, (i) we maintain the number of heads equal to the number of frames in the video clip, thereby generating a unique attention map for each frame; and (ii) we don't need to map values $\mathbf v$ since our goal is feature enhancement rather than encoding.
To achieve the goal in the latter aspect, the $\hat{F}_{sms}$ is first compressed along the channel dimension through Global Adaptive Pooling (GAP) operation to obtain a feature vector with a dimension of $\mathbb{R}^T$, then it continuously passes through an MLP block with two hidden layers and a sigmoid activation function to obtain a weight matrix $\mathbf{W}^E \in \mathbb{R}^{T}$. After that, the $\mathbf{W}^E$ is used to enhance the spatial feature stream $\hat{F}_{sms}$ along the sequence dimension. Furthermore, the residual connection is introduced for training stabilization.
The entire process can be formulated as:
\begin{equation}\label{Eq:Wembed}
\begin{split}
& \mathbf{W}^E = \mathrm{sigmoid}(\textrm{MLP}(\textrm{GAP}(\hat{F}_{sms}))), \\
& O = \mathbf{W}^E \cdot \hat{F}_{sms} + F_{sms},
\end{split}
\end{equation}
where $O \in \mathbb{R}^{T \times C \times H \times W}$ is the refined spatial feature, which is used as the input for the temporal representation learning network DTN.

\subsubsection{Decoupled Temporal Feature Learning (DTN)} \label{sec:decoup_tempo}
As shown in Fig.\ref{fig:pipline}, the DTN mainly consists of Multi-branch Transformer blocks (MTrans). For each block, inspired by~\cite{feichtenhofer2019slowfast, yu2021searching}, we construct multi-branch structures to tackle inputs with different temporal resolutions and feature dimensions accordingly: fast ($\mathbb{R}^{\frac{1}{2}T \times \frac{3}{2}C}$), moderate ($\mathbb{R}^{T \times C}$), and slow ($\mathbb{R}^{\frac{3}{2}T \times \frac{1}{2}C}$), to perceive different motion variations, \ie, the high temporal resolution combined with low channel size and vice versa.
However, Transformer is skilled in global modeling and is insensitive to local fine-grained views~\cite{xiao2021early}. To get around this, we cascade an inception-based local feature extractor (TMS module) before MTrans to perceive fine-grained motion variations.
Formally, for each individual branch $i$, we first sample a subsequence $\hat{O}_i \in \mathbb{R}^{T_i \times C}, i\in \{1, 2, 3, \cdots, n\}$ along the temporal dimension over the feature stream $O$ by a discrete Sampling Layers (SL), written as:
\begin{equation}
\label{Eq:sample}
\begin{small}
\hat{O}_i = \{O_\tau|\tau = \mathcal{R}[\lceil \frac{T}{T_i}\rceil \times t - 1, \lceil \frac{T}{T_i}\rceil \times t], t=1, 2, \dots, T_i \},
\end{small}
\end{equation}
where $\hat{O}_i \subseteq O$, $n$ indicates the number of branches and $\mathcal{R}[a, b]$ represents randomly selecting an integer $x, s.t.~ a \leqslant x \leqslant b$. Then the $\hat{O}_i$ is used as the input to the temporal multi-scale features learning module (TMS).

\noindent\textbf{TMS Module.}\quad 
Similar to SMS, the TMS is composed of a time-centric 3D Inception Module\footnote{The size of the convolution kernel in the spatial dimension is $1 \times 1$.} and a Max Pooling layer, which can be formulated as:
\begin{equation}\label{Eq:local-fine}
F_{tms, i} = \textrm{Maxpool}(\mathrm{InC}_{k\times 1\times 1}(\hat{O}_i, \Theta))
\end{equation}
where $F_{tms, i}$ is local fine-grained temporal features, $\mathrm{InC}_{k\times 1\times 1}(\cdot, \Theta)$ indicates the Inception Module with a kernel size of $k\times 1\times 1$. 
In practice, the temporal receptive field size $k$ is a dynamic value that is dynamically adjusted according to the length of sampled subsequence, \ie, $k=\lceil \sqrt{T_i} \rceil$.
After that, $F_{tms, i}$ is fed into the Transformer block for coarse-grained representation learning.

\noindent\textbf{Transformer block.}\quad To reduce the redundant marginal information in captured temporal features, here we utilize a Transformer structure based on Kvt~\cite{wang2021kvt}. Specifically, for either branch $i$, the temporal features at $l$-th layer can be captured by:
\begin{equation}\label{Eq:global-coarse}
\begin{split}
    & Z^l_i  = \textrm{LN}(\textrm{FFN}(Z^{l-1}_i)) + Z^{l-1}_i, \\
    & Z^{l-1}_i  = \textrm{LN}(\textrm{MSA}_{kNN}(F^l_{tms, i})) + Z^{l-2}_i, \\
\end{split}
\end{equation}
where $\textrm{MSA}_{kNN}(\cdot)$ indicates the $k$-NN multi-head self-attention layer (empirically selecting the top 70\% tokens for training), $\textrm{LN}$ represents Layer Normalization and $\textrm{FFN}$ is Feed Forward Network. 
Furthermore, to avoid the network being biased to one of the sub-branches, we introduce a temperature parameter $\tau$ to control the sharpness of the output distribution of each sub-branch $i$ and impose an individual constraint loss (cross-entropy) on it. Therefore, the output of the DTN network can be given by:
\begin{equation}
\begin{small}
P = \arg\max(\sum_{i=1}^{n}{\textrm{MLP}(Z_{cls, i} / \tau)}), \forall i=1,2,3,\dots, n \\
\end{small}
\end{equation}
where $P$ is the final classification result and $Z_{cls, i}$ is the class token vector embedded in the last Transformer block of $i$-th branch. The $\tau$ follows a cosine schedule from 0.04 to 0.07 during the training. 

\subsubsection{Recoupling Spatiotemporal Representation}\label{sec:recoup}
\label{sec:recoupling}
Considering that spatiotemporal decoupling learning weakens the intrinsic connection between time and space, we thus propose a multi-stage recoupling (MSR) method to reinforce the spatiotemporal correlation through an inner loop optimization mechanism during the training. 
Specifically, we employ a self-distillation loss function $\mathcal{C}_{distill}$ to introduce additional supervision for RCM modules by reversely distilling inter-sequence correlation knowledge from the time domain into the space domain. The distillation process can be written as:
\begin{equation}\label{Eq:distill}
\begin{split}
&\mathcal{C}_{distill} = \frac{1}{L}\sum_{l=1}^{L}{\mathcal{L}_{distill}(\sum_{i=1}^{n}{Z_{cls,i}^l}, \mathcal{F}(\mathbf{W}^{E,l}))}\\
& s.t. \quad \mathcal{L}_{distill}(x, y) =\frac{1}{N} \sum_{i=1}^{N}{KL(x_i/\mathcal{T}-y_i/\mathcal{T})}
\end{split}
\end{equation}
where $L$ is number of RCM modules in DSN, $\mathbf{W}^{E,l}$ is weight matrix (formally defined in Sec.\ref{sec:decoup_spai}) from $l$-th RCM module, $\mathcal{T}$ is the distillation temperature parameter, $N$ is the batch size, $KL$ indicates Kullback-Leibler divergence \cite{hinton2015distilling} and $\mathcal{F}(\cdot)$ indicates linear mapping function, which maps the dimension of $\mathbf{W}^{E, l}$ to align with $Z^l_{cls}$. 
The benefits of the recoupling strategy are manifold. On the one hand, spatiotemporal associations are especially crucial for gesture recognition task because it strongly depends on both location and motion information, \ie, gesture describes a concept of hand variations over time. On the other hand, recouping information can be an effective way to regain common aspects, which can significantly increase the convergence speed of the network. 
Therefore, the entire network is trained by minimizing the following objective:
\begin{equation}
    \mathcal{L}=\mathcal{L}_{entropy} + \lambda_d\mathcal{C}_{distill},
\end{equation}
where $\mathcal{L}_{entropy}$ is the standard classification loss and $\lambda_d$ is the weight coefficient of distillation loss $\mathcal{C}_{distill}$.

\begin{figure}[!t]
\centering
\subfloat[]{\includegraphics[width=0.51\linewidth]{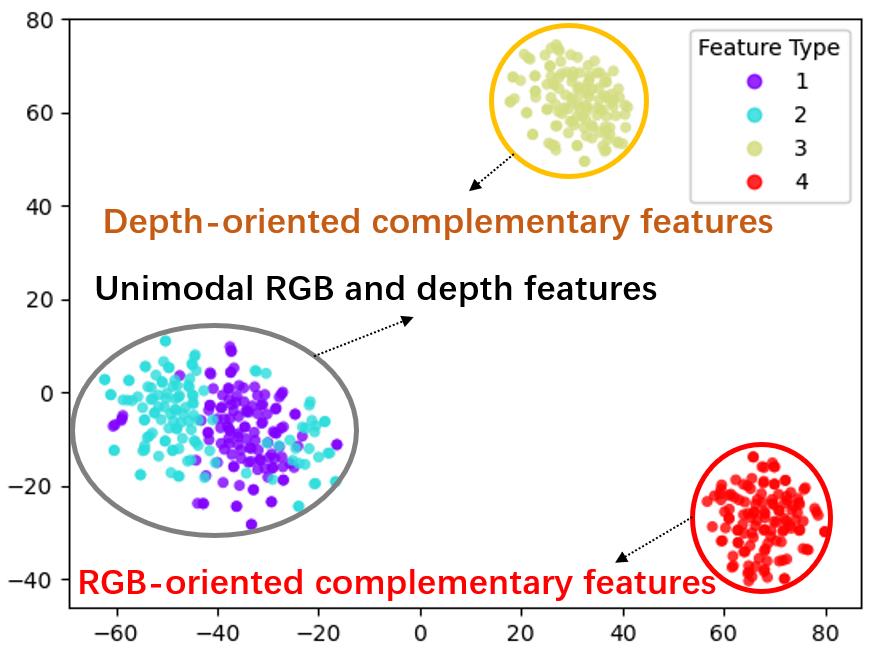}\label{subfig:cluster}}
\hfil
\subfloat[]{\includegraphics[width=0.45\linewidth]{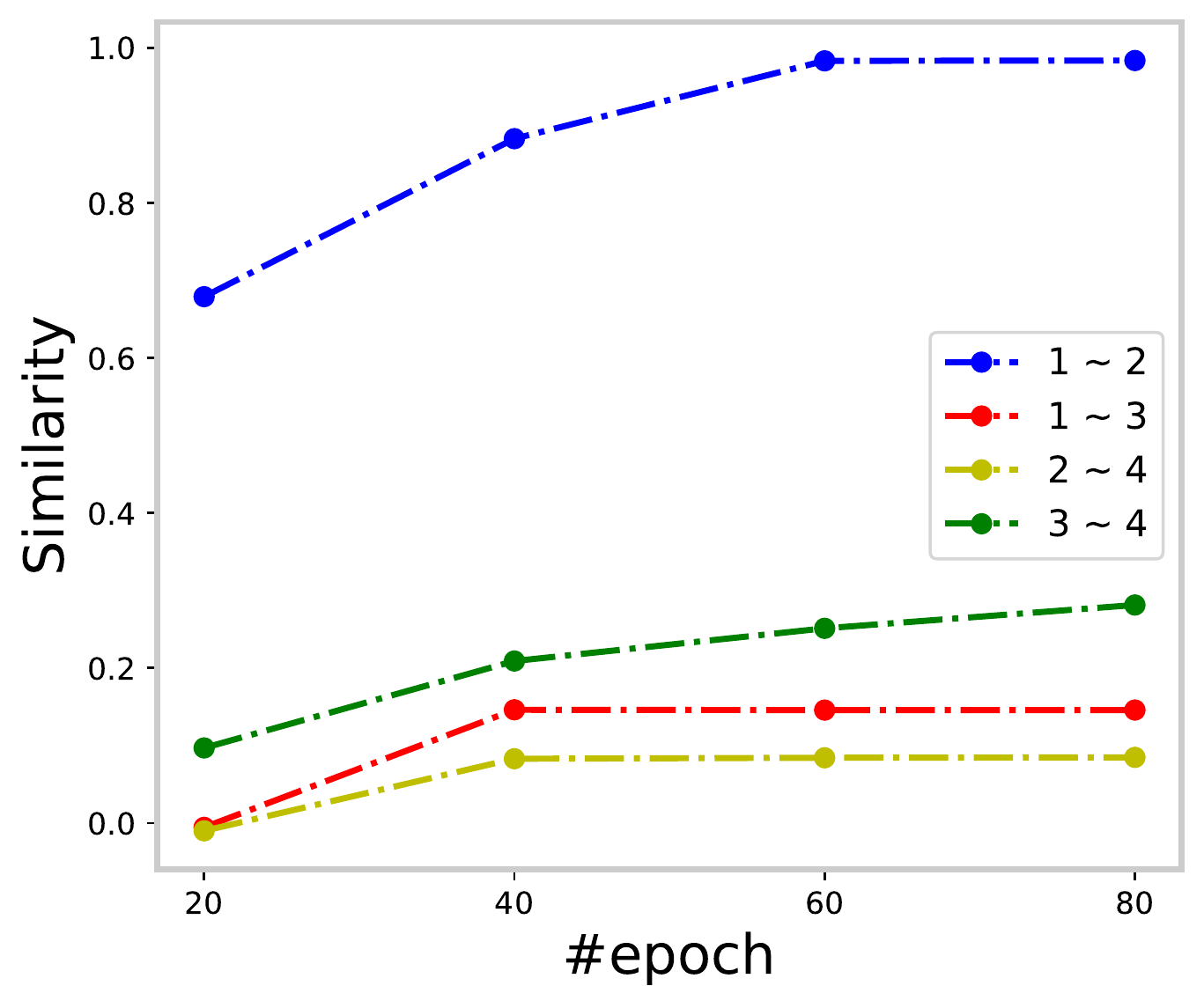}\label{subfig:simil_cuver}}
\caption{(a) \textbf{Visualization of feature distribution dimensional descended by PCA~\cite{PCA2014}.} Feature Type 1/2: the unimodal RGB/depth features output from DTN; Feature Type 3/4: the RGB-/depth-oriented complementary features captured by CFCer. (b) \textbf{Similarity curves of multimodal features.} $\sim$ means the cosine similarity.}
\label{fig:cluster}
\end{figure}
\subsection{Cross-modal Complement Features Learning}
\label{sec:fusion}
The late fusion strategies, including Addition and Multiplication, are simple yet effective multimodal fusion methods that are commonly used in RGB-D-based motion recognition~\cite{li2017largeb, Zhou_Li_Wan_2021, Hu2018Learning}. As we know, the most significant factor affecting late fusion is the similarity between multimodal semantic features~\cite{Hazarika2020misa}, \ie, the distinctive characteristic brings the higher fusion performance. In fact, we hope the classifiers trained from multimodal signals enjoy distinct decision boundaries so that their ensemble can achieve better fusion performance. 
However, we observed a high similarity between multimodal features learned from individual color and depth signals
as shown in Fig.\ref{fig:cluster} (feature types 1 and 2), although they are captured from respective models with different parameters. It means that the two classifiers learned from the unimodal branch may collapse into the same decision surface, suffering from poor fusion results. 
To mitigate this problem, we derive inspiration from the work of Hazarika \etal~\cite{Hazarika2020misa} in the field of Multimodal Sentiment Analysis(MSA), they explore the modality-invariant and modality-specific representations by designing a series of sophisticated fusion loss functions. But our task differs substantially from \cite{Hazarika2020misa}: (i) it is unrealistic to train with multimodal video clips simultaneously when the parameters and input size of the video classification model are considerable; (ii) color and depth data maintain a smaller domain gap than color and audio/text, which leads to the challenge of exploring distinctive complementary information; (iii) we explore the modality-specific representations along spatial to temporal domains.
To this end, inspired by \cite{chen2020bi} and \cite{li2021trear}, we present a novel complementary features learning mechanism as shown in Fig.\ref{fig:CAPFM}, which is lightweight and optimization-friendly, and captures the fine-grained modality-specific aspects by cascading the Complement Feature Catchers (CFCer) along the spatial to temporal domains.

\begin{figure*}[t]
\begin{minipage}[c]{0.58\textwidth}
    \centering
    \captionof{table}{\textbf{Effects of MixUp(MU) and ShuffleMix(SM) on four motion datasets.} We evaluate the results after training 100 epochs. For NTU-RGBD and THU-READ, we report their results on the released CS and CS4 protocols, respectively.
    $\alpha_m \in (0, \infty)$ and $\alpha_s \in (0, \infty)$ mean hyper-parameters of MixUp and ShuffleMix that control the strength of interpolation between sample pairs, herein we consider the most commonly used Beta distribution $Beta(\alpha, \alpha)$. $\rho \in Unif([0, 1])$ represents the probability of selecting ShuffleMix operation.
    }
    \resizebox{1\linewidth}{!}{
    \begin{tabular}{ccccc|cccccccc}
    \toprule
  \multirow{2}{*}{MU} & \multirow{2}{*}{SM} & \multirow{2}{*}{$\alpha_m$} & \multirow{2}{*}{$\alpha_s$} & \multirow{2}{*}{$\rho$} & \multicolumn{2}{c}{\textbf{NvGesture}} & \multicolumn{2}{c}{\textbf{IsoGD}} & \multicolumn{2}{c}{\textbf{NTU(CS)}} & \multicolumn{2}{c}{\textbf{THU(CS4)}}\\
    \cmidrule(r){6-13} 
    & & & & & RGB & Depth & RGB & Depth & RGB & Depth & RGB & Depth\\
    \midrule
    \textcolor{red}{\XSolidBrush} & \textcolor{red}{\XSolidBrush} & - & - & - & 84.13 & 85.71 & 53.98 & 54.45 & 88.99 & 89.61 & 84.28 & 72.22\\
    \midrule
    \textcolor{green}{\CheckmarkBold} & \textcolor{red}{\XSolidBrush}& 0.2 & -& 0.0 & 84.13 & 86.36 & 60.22 & 62.38 & 90.74 & 94.18 &  82.08 & 76.25\\
    \textcolor{green}{\CheckmarkBold} & \textcolor{red}{\XSolidBrush}&0.4 & -& 0.0 & 83.93 & 86.17 & 60.53 & 63.11 & 91.21 & 94.49 & 82.50 & 76.67\\
    \textcolor{green}{\CheckmarkBold} & \textcolor{red}{\XSolidBrush}& 0.6 & -& 0.0 & 82.54 & 85.61 & 60.84 & 63.80 & 91.30 & 94.15 & 84.17 & 76.75\\
    \rowcolor[gray]{0.8} \textcolor{green}{\CheckmarkBold} & \textcolor{red}{\XSolidBrush} & 0.8 &-& 0.0 & 83.93 &  85.61 & 61.61 & 63.88 & 91.56 & 94.55 & 83.86 & 75.83\\
    \textcolor{green}{\CheckmarkBold} & \textcolor{red}{\XSolidBrush}& 1.0 & -& 0.0 & 83.53 & 82.95 & 61.49 & 63.68 & 91.70 & 93.98 & 80.00 &  74.58\\
    
    \midrule
    \rowcolor[gray]{0.8} \textcolor{red}{\XSolidBrush} & \textcolor{green}{\CheckmarkBold} & -& 0.2 & 1.0 & 81.55 & 85.23 & 48.59 & 50.30 & 89.88 & 93.45 & 80.55 & 75.83\\
    \textcolor{red}{\XSolidBrush} & \textcolor{green}{\CheckmarkBold} & -& 0.4 & 1.0 & 79.96 & 85.61 & 48.07 & 49.55 & 89.89 & 93.25 & 78.31 & 75.63\\
    \textcolor{red}{\XSolidBrush} & \textcolor{green}{\CheckmarkBold} & -& 0.6 & 1.0 & 79.16 & 84.09 & 47.83 & 50.22 & 89.80 & 93.13 & 78.62 &  74.33\\
    \textcolor{red}{\XSolidBrush} & \textcolor{green}{\CheckmarkBold} & -& 0.8 & 1.0 & 77.98 & 85.32 & 47.37 & 48.29 & 88.47 & 92.61 & 78.23 & 74.83\\
    \textcolor{red}{\XSolidBrush} & \textcolor{green}{\CheckmarkBold} & -& 1.0 & 1.0 & 78.37 & 82.95 & 46.47 & 41.08 & 88.27 & 92.50 & 77.15 & 74.52\\

    \midrule
    \textcolor{green}{\CheckmarkBold} &  \textcolor{green}{\CheckmarkBold} & 0.8 & 0.2 & 0.1 & 84.79 & 85.80 & 62.87 & 64.26 & 91.62 & 94.62 & 84.29 & 77.00\\
     \textcolor{green}{\CheckmarkBold} &  \textcolor{green}{\CheckmarkBold} & 0.8 & 0.2 & 0.2 & 85.00 & 86.17 & \textbf{63.68} & \textbf{64.62} & 92.23 & \textbf{94.86} & 84.33 & 77.08\\
     \textcolor{green}{\CheckmarkBold} &  \textcolor{green}{\CheckmarkBold} & 0.8 & 0.2 & 0.3 & 85.21 & 86.93 & 62.92 & 64.52 & \textbf{92.24} & 94.71 & 84.38 & 77.83\\
     \textcolor{green}{\CheckmarkBold} &  \textcolor{green}{\CheckmarkBold} & 0.8 & 0.2 & 0.4 & 84.95 & 86.36 & 61.09 & 63.44 & 91.49 & 94.71 & 85.89 & 77.42\\
     \textcolor{green}{\CheckmarkBold} &  \textcolor{green}{\CheckmarkBold} & 0.8 & 0.2 & 0.5 & \textbf{86.25} & \textbf{88.04} & 61.29 & 63.59 & 91.50 & 94.77 & \textbf{87.50} & \textbf{78.33}\\
     \midrule
     \multicolumn{5}{c|}{$\bigtriangleup$} & $\uparrow 2.1$ & $\uparrow2.3$ & $\uparrow 9.7$ & $\uparrow10.2$ & $\uparrow3.3$ & $\uparrow5.3$ & $\uparrow3.2$ & $\uparrow6.1$\\
    \bottomrule
    \end{tabular}
    }
    \label{tab:mix_ratio}
\end{minipage}%
 \hfill%
\begin{minipage}[c]{0.4\textwidth}
  \centering
   \subfloat[Compound gestures with multiple motion trajectories.]{\includegraphics[width=0.9\linewidth]{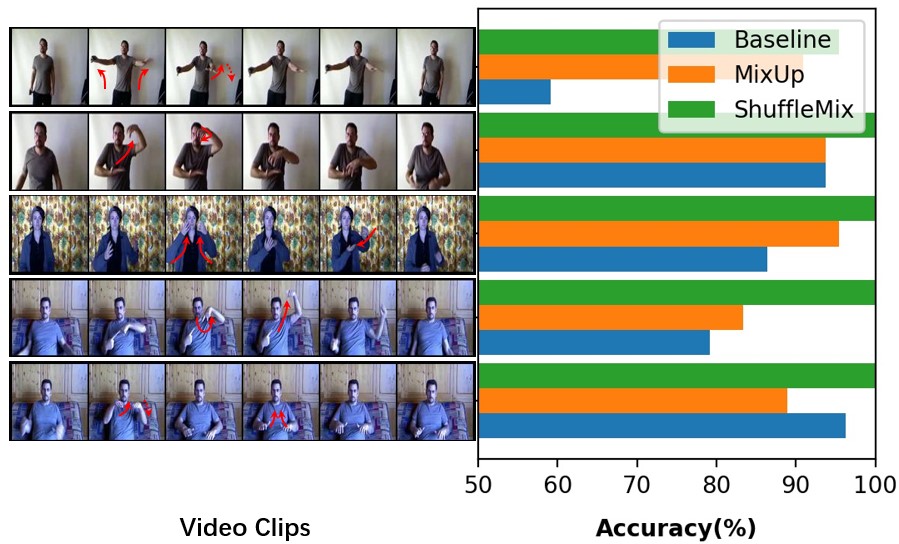}\label{subfig:shuffle_effect}}
    \hfil
    \subfloat[Simple gestures with single motion trajectory.]{\includegraphics[width=0.9\linewidth]{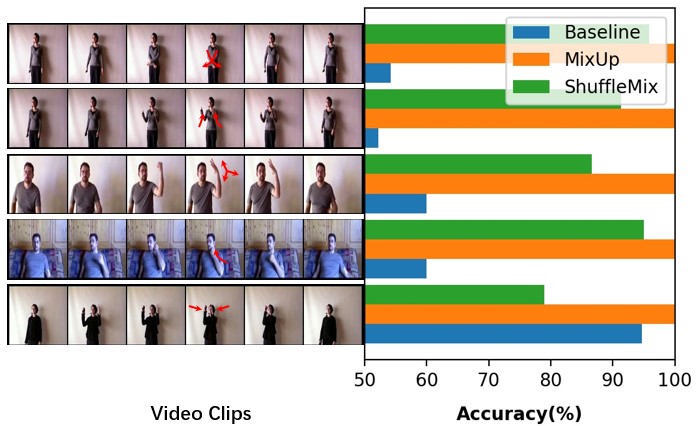} \label{subfig:mixup_effect}}
   \captionof{figure}{\textbf{Effect of MixUp and ShuffleMix on IsoGD.} (a)/(b): The top-5 classes that were significantly improved by ShuffleMix/MixUp.
}
   \label{fig:class_idx}
\end{minipage}%
\end{figure*}

\noindent\textbf{CFCer.}\quad As illustrated in Fig.\ref{subfig:cfc}, the CFCer contains a feature reinforcement component and mutual-attention mechanism. 
Taking the color-oriented complementary features learning as an example. Let $O_{r}$ and $O_{d}$ be the unimodal feature maps learned from color and depth data, respectively. They are first concatenated along the channel dimension and normalized by a Layer Normalization (LN). To highlight the most descriptive and modality-specific features, an MLP network with multiple hidden layers is trained together to calculate the channel-wise attention score, followed by sigmoid re-scaling of the weight values:
\begin{equation}
\begin{split}
    &\tilde{O}_{r} = \textrm{sigmoid}(\textrm{MLP}(\textrm{LN}([O_{r}||O_{d}]))) \odot O_{r},
\end{split}
\end{equation}
where $||$ represents concatenation, $\odot$ denotes element-wise multiplication, and $\tilde{O}_{r}$ is enhanced color feature maps. Then, we mine color-oriented complementary features from multimodal information using a mutual-attention structure, which is designed to retrieve information related to color (depth) cues from the depth (color) stream. 
Specifically, first, we define a set of learnable parameter matrices $ \mathbf W^Q$, $\mathbf W^K$ and $\mathbf W^V$ to obtain the query vector from the depth modality, and the key and value vectors from the color modality:
\begin{equation}
      \mathbf{q}_d=\tilde{O}_{d} \cdot \mathbf W^Q,\quad \mathbf k_r= \tilde{O}_{r} \cdot \mathbf W^K, \quad \mathbf v_r= \tilde{O}_{r} \cdot \mathbf W^V
\end{equation}
where $\tilde{O}_{d}$ means the enhanced depth feature maps. Then, $\mathbf{q}_d$ and $\mathbf k_r$ are utilized to calculate the attention score:
\begin{equation}
        \mathbf{A}_r = \textrm{softmax}(\frac{\mathbf{q}_d \cdot \mathbf{k}^T_r}{\sqrt{d}})
\end{equation}
where $d=64$ is the scaling factor. Subsequently, the color-oriented complementary features  $F_{comp, r}$ can be calculated by:
\begin{equation}
    F_{comp, r} = \mathbf{v}_r \cdot \mathbf{A}_r.
\end{equation}
As can be seen in Fig.\ref{subfig:cluster} (feature types 3 and 4) and Fig.\ref{subfig:simil_cuver}, the color- and depth-oriented complementary features captured by CFCer maintain lower similarity compared to unimodal features, highlighting the effectiveness of CFCer in learning complex patterns and features from multimodal data.

In practice, before we optimize the CFCers, we initialize the corresponding unimodal network branch with the parameters trained on the color/depth data and freeze them throughout training. 
In more detail, as illustrated in Fig.\ref{fig:CAPFM}, we disentangle the task of spatiotemporal fine-grained complementary representations learning into 2 steps: (i) capturing the appearance-level complimentary correspondences from the spatial features by deploying two-layer CFCer in shallow layers; (ii) concatenating them with the semantic class token embedding from the corresponding DTN network and learning semantic-level complementary representations by deploying four-layer CFCer in the deep layers.
After that, the captured complementary features are fed into their respective classifiers to obtain distinct predictions. They serve as an auxiliary fusion stream, together with the unimodal predictions, to improve the late fusion. In particular, CFCer extracts the modality-specific aspects in a more straightforward manner that is only supervised by a classification loss. In other words, CFCer learns complementary features by fine-grained retrieval of classification cues implicit in the peer modality, enabling it to achieve the goal without the need for elaborate adversarial loss functions.

\section{Experiments}
In this section, we first give details for benchmark datasets (Sec.\ref{sec:dataset}) and experimental setup (Sec.\ref{sec:setup}). Then, we thoroughly evaluate the effectiveness of each proposed technique (Sec.\ref{sec:study}). Finally, we compare our results with state-of-the-art methods on four benchmark RGB-D motion datasets (Sec.\ref{sec:sota}).

\subsection{Datasets} \label{sec:dataset}
We evaluate our method on four public RGB-D motion datasets: NTU RGB-D \cite{shahroudy2016ntu},  NvGesture \cite{molchanov2016online}, Chalearn IsoGD ~\cite{wan2016chalearn, 9172121} and THU-READ \cite{tang2018multi} datasets.
The\textbf{ NTU RGB-D} is a large-scale human action dataset containing more than 56,000 multi-view RGB-D videos with 4 million frames that include 60 actions performed by 40 subjects. This dataset is challenging due to large intra-class and viewpoint variations. 
In addition, skeleton information is also provided in this dataset. This dataset releases two validation protocols namely Cross Subject (CS) and Cross View (CV).
The \textbf{NvGesture} dataset focuses on touchless driver controlling. It contains 1532 dynamic gestures fallen into 25 classes involving RGB-D videos and a pair of stereo-IR streams. This dataset is divided into training and testing subsets with a proportion of 7:3, namely 1050 samples for training and 482 for testing. 
The \textbf{Chalearn IsoGD} dataset contains 47,933 RGB-D gesture videos divided into 249 kinds of gestures and is performed by 21 individuals. It has three subsets of training, validation, and test set,  containing 35878, 5784, and 6271 samples respectively. Samples in the three subsets are exclusive. It is also used as the benchmark for two rounds of the Chalearn LAP large-scale isolated gesture recognition challenge. Similar to most of the recent works \cite{wang2020hybrid, yu2020multi, Zhou_Li_Wan_2021, Zhou_2022_CVPR}, we also evaluate the experimental results on the validation set for this dataset.
The \textbf{THU-READ} is the first-person view RGB-D dataset that consists of 1920 videos with 40 different actions performed by 8 subjects. And each subject repeats each action 3 times. Similar to other works, we adopt the released leave-one-split-out cross-validation protocol in our experiments, which divides the 8 subjects into 4 groups and uses 3 splits for training and the rest for testing.

\subsection{Implementation Details} \label{sec:setup}
The proposed method is implemented with Pytorch. 
The input sequences are random/center cropped into $224\times224$ during training/inference.
We employ SGD as the optimizer with a weight decay of 0.0003 and momentum of 0.9. The learning rate is linearly ramped up to 0.01 during the first 5 epochs and then decayed with a cosine schedule~\cite{loshchilov2016sgdr}. The training lasts for 100 epochs. Similar to \cite{Zhou_Li_Wan_2021,yu2021searching,Zhou_2022_CVPR}, all of our experiments except NTU-RGBD are pre-trained on 20BN Jester V1 dataset~\cite{materzynska2019jester}.  In addition, three SMS modules are configured in the DSN, each followed by an RCM module. In the DTN, we configure three individual branches accordingly: slow, moderate, and fast. Each unimodal branch is configured with a TMS module and six MTrans blocks. For ShuffleMix+, we set the hyper-parameters \{$\alpha_m$, $\alpha_s$, $\rho$\} as \{0.8, 0.2, 0.5\} for NTU-RGBD and IsoGD, while \{0.8, 0.2, 0.2\} for NvGesture and THU-READ, according to the ablation study in Sec.\ref{sec:shuffle+}.
We refer to this setting as the basic configuration of our network unless otherwise specified. 
\begin{table}[!t]
  \caption{\textbf{The effects of multi-stage recoupling (MSR) strategy.} Where MSR-$i$ represents the number of RCM modules embedded in the DSN. MSR-1 means the single-stage recoupling strategy.}
  \label{tab:msr}
  \centering
  \begin{tabular}{cc|cc}
    \toprule
     \textbf{method} & \#\textbf{frames} & \textbf{NTU-RGBD(CS)} & \textbf{NvGesture}\\
    \midrule
    MSR-1 & 16 & 90.81 & 79.91\\
    MSR-2 & 16 & 91.08 & 81.14\\
    \rowcolor[gray]{0.8} MSR-3 & 16 & \textbf{91.37} &  \textbf{82.29}\\
    \midrule
    MSR-1 & 32 & 91.46 & 84.52 \\
    MSR-2 & 32 & 91.86 & 85.71\\
    \rowcolor[gray]{0.8} MSR-3 & 32 & \textbf{92.24} & \textbf{86.25} \\
    \bottomrule
  \end{tabular}
\end{table}

\subsection{Ablation Study}\label{sec:study}
In this section, all of our investigations are based on the RGB data except Sec.\ref{sec:exp_fusion} and Sec.\ref{sec:contribute}.

\begin{table}[!t]
    \centering
    \caption{\textbf{Effect of different implementations of ShuffleMix+.} Where dis-mix and con-mix represent global-based discrete mixture and local-based continuous mixture. The results are reported on NTU-RGBD(CS), IsoGD, and NvGesture datasets under the 32-frame rate.}
    \label{tab:shuffimpl}
    \resizebox{1\linewidth}{!}{
    \begin{tabular}{cccc|ccc}
    \toprule
     \textbf{batch-wise} & \textbf{pair-wise} & \textbf{dis-mix} & \textbf{con-mix} & \textbf{NTU} & \textbf{Iso} & \textbf{Nv} \\
    \midrule
    \rowcolor[gray]{0.8} \textcolor{green}{\CheckmarkBold} & & \textcolor{green}{\CheckmarkBold}& & \textbf{92.24} & \textbf{63.68} & 86.25\\
    \textcolor{green}{\CheckmarkBold} & &  & \textcolor{green}{\CheckmarkBold} & 92.19 & 63.18 & \textbf{86.88}\\
    & \textcolor{green}{\CheckmarkBold} & \textcolor{green}{\CheckmarkBold}& & 91.41 & 63.10 & 86.04\\
    \bottomrule
    \end{tabular}
    }
\end{table}
\subsubsection{Effect of Spatiotemporal Regularization} \label{sec:shuffle+}
To highlight the importance of spatiotemporal regularization for motion recognition, we conduct extensive experiments on four motion datasets with RGB-D modalities.
As illustrated in Table.\ref{tab:mix_ratio}, MixUp can significantly improves motion recognition on large-scale datasets under the appropriate interpolation strength. Taking the RGB modality as example, compared with baseline (\ie, line 1 in Table.\ref{tab:mix_ratio}), MixUp gains 7.63\% on IsoGD when $\alpha_m=0.8$ and gains 2.71\% on NTU-RGBD when $\alpha_m=1.0$, which indicates the advantages of spatial regularization. However, this strategy seems to be invalidated on two small-scale motion datasets, \eg, no gains for NvGesture, and even inferior performance for THU-READ, which means that it still remains challenging to learn spatiotemporal consistency features from small-scale data settings only through spatial-level regularization. In other words, mixing information only at the spatial level is insensitive to temporal evolution, which limits the regularization effect of MixUp in the video domain.
Different from MixUp, ShuffleMix provides additional temporal regularization for video representation learning. As can be seen, when combined it with MixUp, they significantly improve performance on both color and depth data especially for small-scale datasets, \ie, improving $\uparrow 2.1\%$ and  $\uparrow 2.3\%$ on NvGesture, $\uparrow 3.2\%$ and  $\uparrow 6.1\%$ on THU-READ when $\alpha_m=0.8$, $\alpha_s=0.2$, $\rho=0.5$ compared with baseline.

In addition, we interestingly find that ShuffleMix improves a lot on more complex compound gestures as shown in Fig.~\ref{subfig:shuffle_effect}, while MixUp is more friendly to relatively simple single-track gestures as shown in Fig.~\ref{subfig:mixup_effect}. This shows that ShuffleMix has stronger fine-grained motion perception capabilities than MixUp. Therefore, this is why their combination can bring about significant performance gains in motion recognition compared to their single use.

\begin{figure}[!t]
\centering
\subfloat[Results on NTU-RGBD(CS).]{\includegraphics[width=0.5\linewidth]{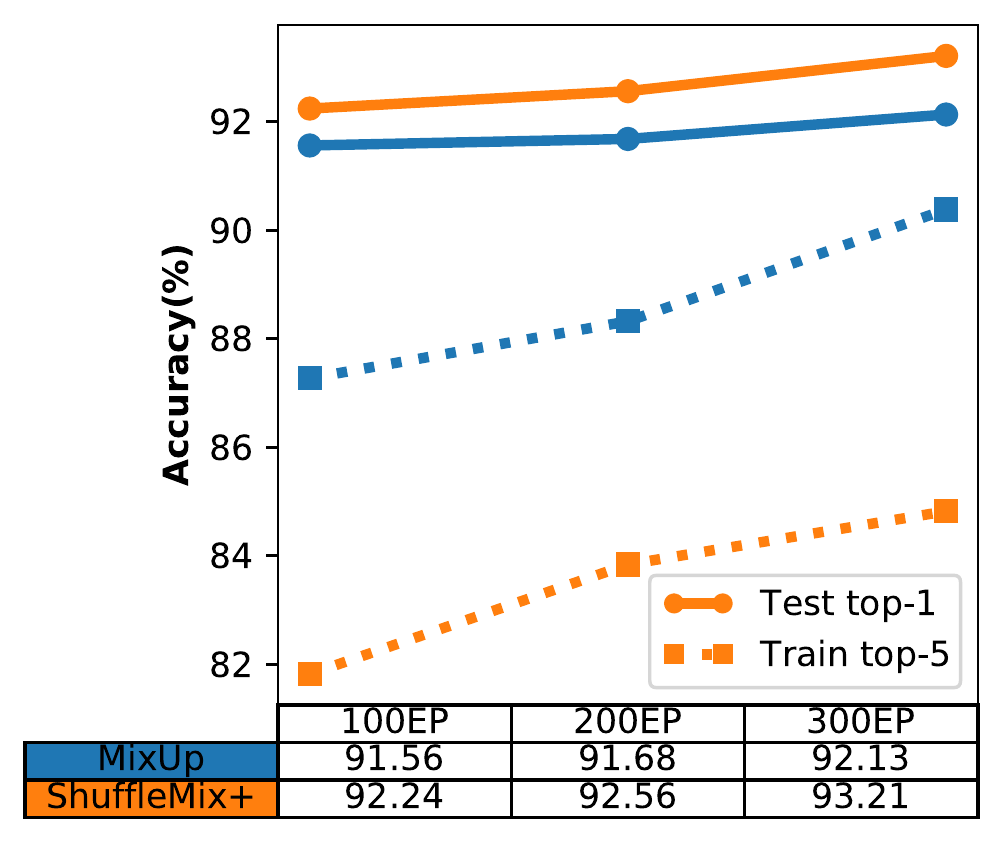}%
\label{subfig:long_regim_ntu}}
\hfil
\subfloat[Results on NvGesture.]{\includegraphics[width=0.5\linewidth]{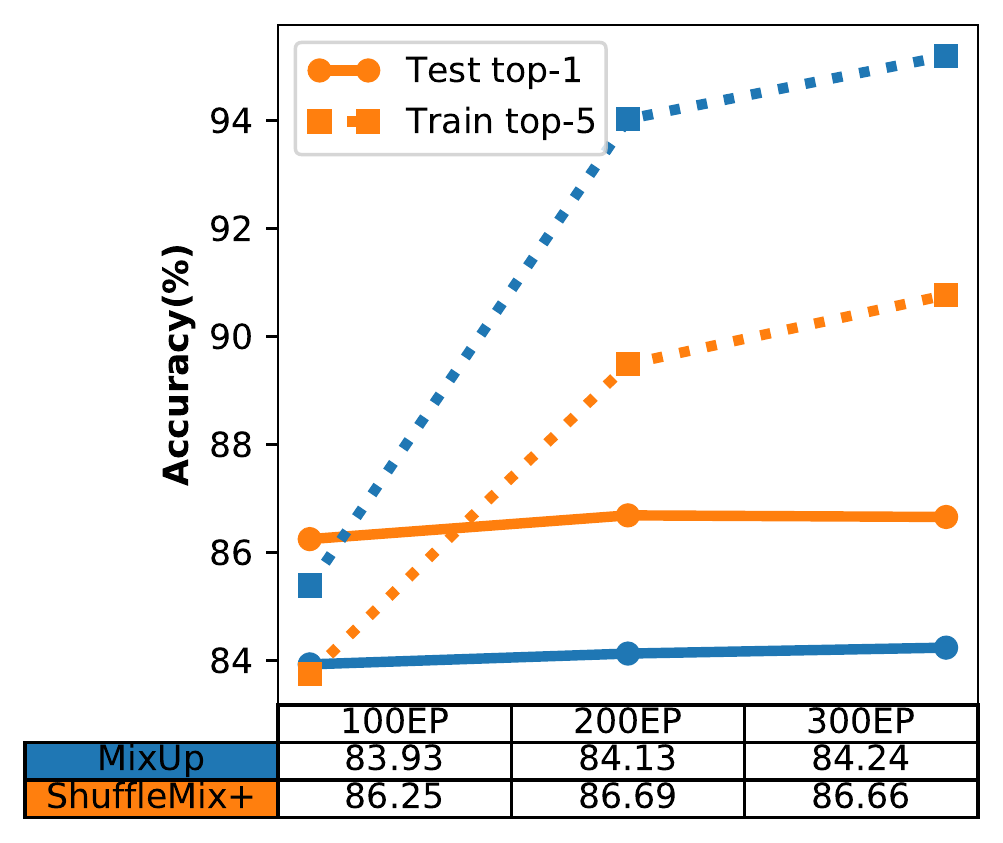}%
\label{subfig:long_regim_nv}}
\caption{\textbf{Effect of longer training regimes.} The table below the figure shows the top-1 accuracy on the testing set, and EP means epoch. Test top-1 and Train top-5 mean top-1 accuracy evaluated on the testing set and top-5 accuracy evaluated on the training set, respectively. The results are reported on NTU-RGBD(CS) and NvGesture datasets under the 32-frame rate.}
\label{fig:long_regim}
\end{figure}
\subsubsection{Effect of Multi-stage Recoupling Learning}
In this experiment, we demonstrate the effectiveness of the Multi-Stage Recoupling (MSR) strategy on the NvGesture and NTU-RGBD datasets under the different frame rates, because they come from different domains as well as have different scales. As illustrated in Table.\ref{tab:msr}, under the same frame rate, the more the recoupling module it has, the better performance it will be.
The reason might lie in the following aspects. On the one hand, multi-stage spatial feature enhancement helps to select discriminative feature maps and improve the robustness of the model. On the other hand, MSR encourages to further enhancing the space-time connection for compact spatiotemporal representation learning.

\subsubsection{Effect of Different Implementations of ShuffleMix+}
\label{sec:diffshuff}
In this experiment, we investigate two proposed video mixing strategies: a local-based continuous mixing method and a global-based discrete mixing method. Additionally, two ShuffleMix+ implementations, namely batch-wise and pair-wise, are also explored, where the former refers to randomly operating ShuffleMix+ across mini-batches, and the latter refers to randomly operating within a mini-batch. 
Considering that the IsoGD involves a wide range of gestures from different domains and scenarios, the NvGesture contains more fine-grained motions, and NTU-RGBD contains actions from multiple perspectives. We, therefore, conduct experiments on these three datasets to demonstrate universality.

\begin{figure}[!t]
  \centering
   \subfloat[Addition/Multiplication]{\includegraphics[width=0.4\linewidth]{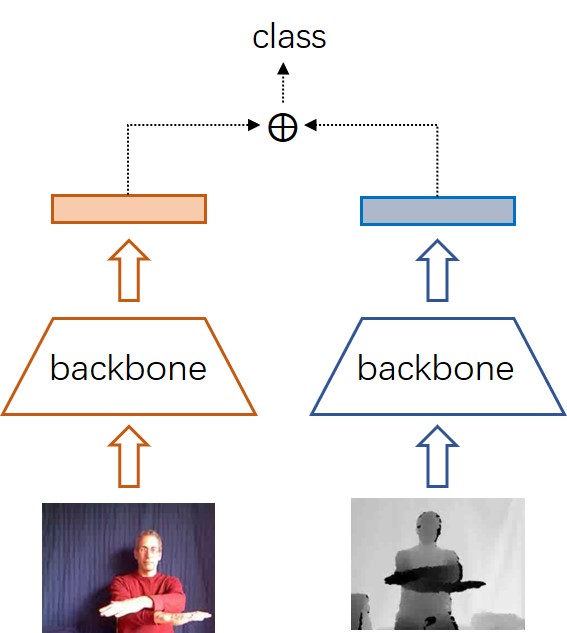}\label{subfig:fusion_add}}
    \hfil
    \subfloat[SFN]{\includegraphics[width=0.4\linewidth]{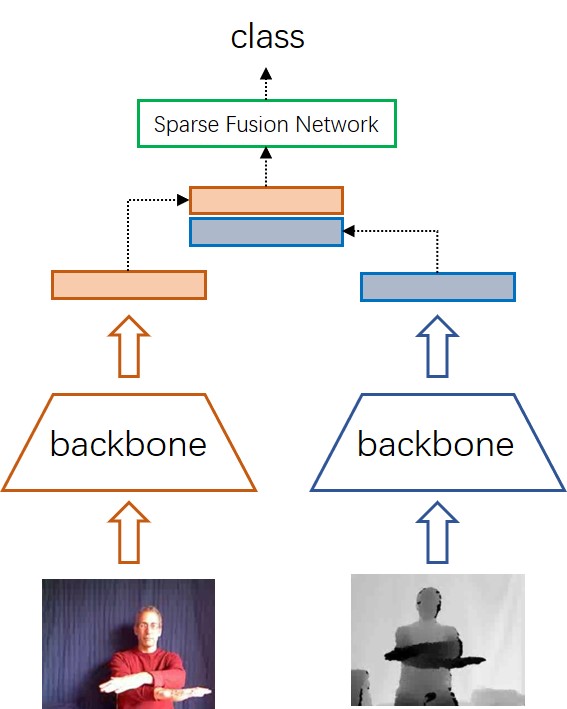} \label{subfig:fusion_sfn}}
    \hfil
    \subfloat[CAPF]{\includegraphics[width=0.4\linewidth]{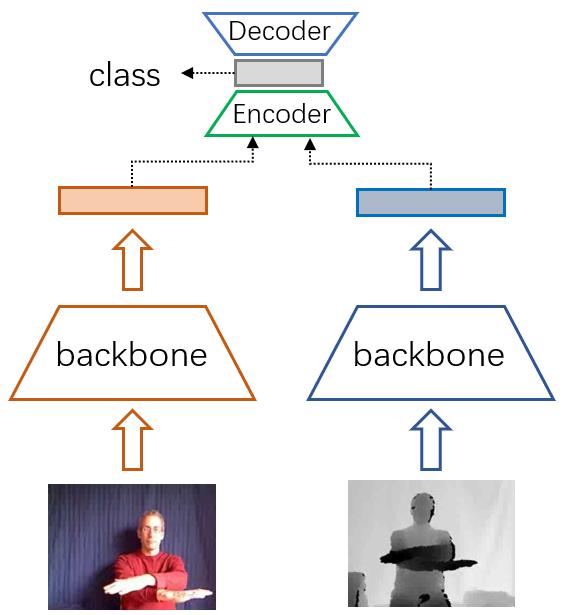} \label{subfig:fusion_capf}}
    \hfil
    \subfloat[CFCer]{\includegraphics[width=0.4\linewidth]{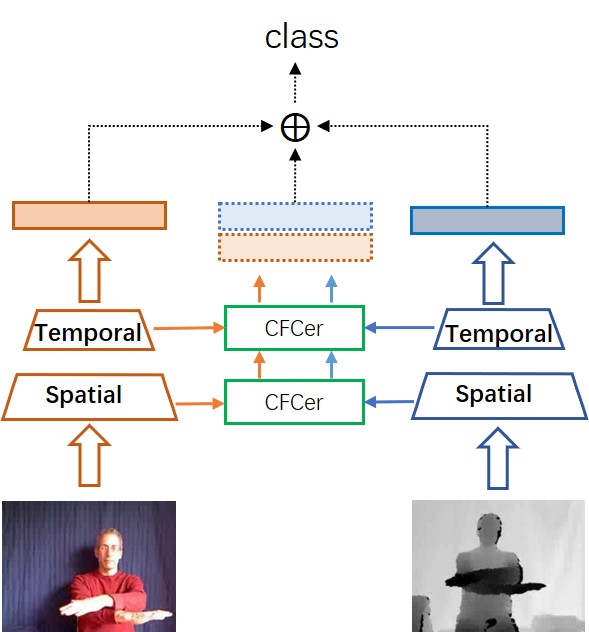} \label{subfig:fusion_cfcer}}
   \caption{\textbf{Four commonly used late fusion methods for RGB-D motion recognition.}
}
   \label{fig:fusion_strategies}
\end{figure}%
As shown in Table.\ref{tab:shuffimpl}, we observe that the discrete mixture method outperforms the continuous mixture method on both NTU-RGBD and IsoGD datasets, while the latter is more friendly to NvGesture. This is because NvGesture contains more fine-grained gestures that are more concerned with local continuity of motion. Considering that the performance gap between these two mixture methods is not obvious, we thus uniformly adopt a discrete mixture strategy in this paper for convenience.
Furthermore, for the two ShuffleMix+ implementations, batch-wise and pair-wise, our experiments show that the former improves more even though they are formally equivalent.

\begin{table}[!t]
  \caption{\textbf{Effect of recoupling loss weighting coefficients on performance.}}
  \label{tab:loss_weitht}
  \centering
  \begin{tabular}{cc|cc}
    \toprule
     $\lambda_d$ & \textbf{\#frames} & \textbf{NTU-RGBD(CS)} & \textbf{NvGesture}\\
    \midrule
    0.0 & 32 & 91.56 & 85.04 \\
    0.1 & 32 & 91.75 & 85.46 \\
    \rowcolor[gray]{0.8} 0.2 & 32 & \textbf{92.24} & \textbf{86.25}\\
    0.3 & 32 & 91.87 & 85.83\\
    \bottomrule
  \end{tabular}
\end{table}
\subsubsection{Effect of ShuffleMix+ under Longer Training Regimes}
As illustrated in Fig.\ref{fig:long_regim}, a longer training schedule continuously improves performance, and the phenomenon is more pronounced on large-scale datasets. In fact, we find that ShuffleMix+ achieves significant improvement on NTU-RGBD ($\uparrow$1\% when training 300 epochs) over longer training regimes than using MixUp alone, which means that the ShuffleMix+ further enlarges the knowledge capacity of training data space. Meanwhile, we find that models trained with ShuffleMix+ had a lower top-5 accuracy on the training set than MixUp (dotted line in Fig.\ref{fig:long_regim}), implying that ShuffleMix+ encourages stronger regularization to further alleviate the overfitting problem faced by the model. 
Considering the fairness and time efficiency, all our experiments are trained for only 100 epochs.

\subsubsection{Anatomy of Recoupling Loss Weight}
As mentioned before, the multi-stage recoupling strategy is proposed to reconstruct the spatiotemporal dependencies from shallow to deep layers, which is implemented by the self-distillation strategy. Where the weight coefficient $\lambda_d$ controls the strength of spatiotemporal connections. As can be seen in Table.\ref{tab:loss_weitht}, recoupling is more friendly to gesture recognition ($\uparrow$1.2\% when $\lambda_d$=0.2), which suggests that the gesture is more dependent on the intrinsic relationship between space and time than action.

\begin{table}[!t]
  \caption{\textbf{Effect of different late fusion strategies.}}
  \label{tab:fusion}
  \centering
  \resizebox{1\linewidth}{!}{
  \begin{tabular}{lc|cccc}
    \toprule
     \textbf{fusion} & \#\textbf{frames} & \textbf{NTU(CS)} & \textbf{Nv} & \textbf{Iso} & \textbf{THU(CS4)}\\
    \midrule
    Addition & 32 & 95.44 & 89.03 & 72.23 & 88.33\\
    Multiplication & 32 & 95.13 & 88.94 & 72.01 & 86.25 \\
    SFN\cite{narayana2018gesture} & 32 & 95.52 & 89.18 & 72.24 & 88.35\\
    CAPF\cite{Zhou_2022_CVPR}& 32 & 95.58 &  89.43 & 72.34 & 88.49\\
    \rowcolor[gray]{0.8} \textbf{CFCer} & 32 & \textbf{95.87} & \textbf{90.04} & \textbf{72.61} & \textbf{89.10} \\
    \bottomrule
  \end{tabular}
  }
\end{table}
\subsubsection{Effect of Fusion Strategy}
\label{sec:exp_fusion}
In this section, as illustrated in Fig.\ref{fig:fusion_strategies}, four commonly used late fusion methods for RGB-D motion recognition are presented for comparison. Among them, \textit{Addition} and \textit{Multiplication} simply add/multiply the predicted scores from the unimodal branches for classification. SFN\cite{narayana2018gesture} introduces a sparse fusion network to learn how much weight to assign to a channel. CAPF\cite{Zhou_2022_CVPR} proposes a lightweight fusion mechanism with an Encoder and Decoder, and trains it through a multi-loss joint optimization strategy. Our CFCer improves the late fusion by exploring the modality-specific complementary features along the spatial to temporal domains.

For \textit{Addition} and \textit{Multiplication}, we simply add/multiply the predicted scores from the unimodal branches,  while for SFN\cite{narayana2018gesture}, CAPF\cite{Zhou_2022_CVPR} and CFCer, we first train the unimodal network branch on corresponding color/depth data, and then freeze their parameters and fit fusion layers with a few training schedules by cross-entropy loss.
As can be seen in Table.\ref{tab:fusion}, the CFCer-based fusion strategy performs best than other late fusion methods. This is because CFCer separates modality-specific complementary features from multimodal semantic information as an auxiliary fusion stream, inheriting the inherent semantic content and preserving the distinction from unimodal representations, which is ignored by other fusion strategies. 
\begin{table}[!t]
  \caption{\textbf{Effects of configuring different numbers of CFCer in the spatial and temporal domains.}}
  \label{tab:cfcer}
  \centering
  \begin{tabular}{cccc|cc}
    \toprule
     \textbf{spatial} & \textbf{temporal} & \textbf{\#params} &\textbf{\#frames} & \textbf{NTU(CS)} & \textbf{Nv}\\
    \midrule
    2 & 2 & 10.5M & 32  & 95.63 & 89.51\\
    \rowcolor[gray]{0.8}2 & 4 & 13.4M & 32 & 95.87 & 90.04  \\
    4 & 2 & 13.4M & 32 &  95.70  &  89.63 \\
    2 & 6 & 16.3M & 32 & \textbf{95.90}  &  \textbf{90.17} \\
    \bottomrule
  \end{tabular}
\end{table}
\begin{table}[!htp]
  \caption{The contributions of the individual (RGB/depth) and fused (RGB-D) channels.
  }
  \label{tab:channel}
  \centering
  \resizebox{1\linewidth}{!}{
  \begin{tabular}{c|cccc}
    \toprule
     \textbf{modality} & \textbf{IsoGD} & \textbf{NTU-RGBD(CS)} & \textbf{NvGesture} & \textbf{THU-READ}\\
    \midrule
    RGB & 64.4 & 92.9 & 89.8 & 82.9\\
    Depth & 65.5 & 95.0 & 91.2 & 80.9\\
    RGB-D & 72.7 & 96.2 & 92.0 & 90.0\\
    \bottomrule
  \end{tabular}
  }
\end{table}
\begin{figure*}
    \centering
    \includegraphics[width=1.0\linewidth]{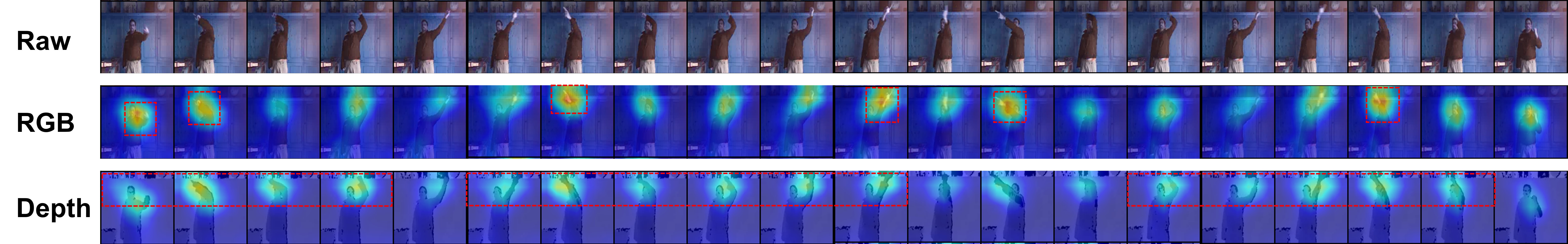}
    \caption{The visualization of activation maps for RGB and depth modalities. The RGB modality perceives limbs in motion, while the depth modality perceives the movement of the limbs.}
    \label{fig:visualize}
\end{figure*}
\begin{table}[!t]
  \centering
    \caption{Comparison of the SOTA methods on the NTU-RGBD.}
  \label{tab:ntu}
  \resizebox{1\linewidth}{!}{
  \begin{tabular}{@{}lcccc@{}}
    \toprule
    \textbf{method} & \textbf{modality} & \#\textbf{frames} & \textbf{CS(\%)} & \textbf{CV(\%)}\\
    \midrule
    Directed-GNN~\cite{shi2019skeleton} & Skeleton & 300 & 89.9 & 96.1\\
    Shift-GCN~\cite{2020Skeleton} & Skeleton & 300 & 90.7 & 96.5\\
    DC-GCN+ADG ~\cite{cheng2020decoupling} & Skeleton & 300 & 90.8 & 96.6\\
    CTR-GCN ~\cite{chen2021channel} & Skeleton & 64 & \textbf{92.4} & \textbf{96.8}\\
    \midrule
    Chained Multi-stream ~\cite{2017Chained} & RGB & 16 & 80.8 & -\\
    HybridNet~\cite{wang2020hybrid}& RGB & 32 & 86.5 & 88.5 \\
    Decouple+Recouple~\cite{Zhou_2022_CVPR} & RGB & 64 & 90.3 & 95.4\\
    \textbf{UMDR} & RGB & 16 & 92.0 & 95.3\\
    \textbf{UMDR} & RGB & 32 & 92.2 & 95.8 \\
    \textbf{UMDR} & RGB & 64 & \textbf{92.9} & \textbf{96.3}\\
    \midrule
    SLTEP~\cite{2017The} & Depth & - & 58.2 & -\\
    DynamicMaps+CNN ~\cite{2018Depth} & Depth & - & 87.1 & 84.2\\
    HybridNet~\cite{wang2020hybrid} & Depth & 32 & 87.7 & 87.4 \\
    Decouple+Recouple~\cite{Zhou_2022_CVPR} & Depth & 64 & 92.7 & 96.2\\
    \textbf{UMDR} & Depth & 16 & 94.5 & 95.4\\
    \textbf{UMDR} & Depth & 32 & 94.8 & 95.9\\
    \textbf{UMDR} & Depth & 64 & \textbf{95.0} & \textbf{96.5}\\ 
    \midrule
     Deep Bilinear ~\cite{Hu_2018_ECCV} & RGB-D-Pose & 16 & 85.4 & 90.7\\
    P4Transformer ~\cite{Fan_2021_CVPR} & Point clouds & - & 90.2 & 96.4\\
    MMTM ~\cite{Joze_2020_CVPR} & RGB+Pose & 32 & 91.9 & -\\
    PoseC3D ~\cite{Duan_2022_CVPR} & RGB+Pose & 16+32 & 94.1 & 97.1\\
    \midrule
    c-ConvNet~\cite{wang2018cooperative} & RGB-D & 4 & 86.4 & 89.1\\
    HybridNet~\cite{wang2020hybrid} & RGB-D & 32 & 89.5 & 91.7 \\
    Decouple+Recouple~\cite{Zhou_2022_CVPR} & RGB-D & 64 & 94.2 & 97.3\\
    \textbf{UMDR (CFCer)} & RGB-D & 16 & 95.6 & 97.5\\
    \textbf{UMDR (CFCer)} & RGB-D & 32 & 95.9 & 97.8\\
    \textbf{UMDR (CFCer)} & RGB-D & 64 & \textbf{96.2} & \textbf{98.0}\\
    
    \bottomrule
  \end{tabular}
  }
\end{table}
In addition, we investigate the effects of configuring different numbers of CFCer in the spatial and temporal domains in Table.\ref{tab:cfcer}. As mentioned before, CFCer mainly captures appearance-level complementary correspondences in the spatial domain and semantic-level complementary correspondences in the temporal domain. As can be seen, the more the CFCer in the temporal domain, the better performance it will be. It implies that complementary features rich in semantic information dominate the fusion performance. Considering the trade-off between performance and parameters, we configure 2 and 4 CFCers for the spatial and temporal domains, respectively.
\subsubsection{Contributions of RGB and Depth Channels}\label{sec:contribute}
Table.\ref{tab:channel} quantitatively analyzes the contributions of the individual (RGB/depth) and fused (RGB-D) channels. As can be seen, in most cases, the contribution of the depth channel is dominant, benefiting from its strong anti-background noise ability and motion perception. In fact, as shown in Fig.\ref{fig:visualize} (red dashed box), the color channel emphasizes the perception of space because it is based on the visual information that is available in the environment, such as color, texture, and shape. The depth channel emphasizes the perception of movement because it is based on the relative distances or positions between objects in the environment, which can be used to determine the direction and speed of movement. Therefore, for actions that rely less on movement, a depth channel may not be as effective. For instance, in the THU-READ dataset, many of the actions can be classified with just a few static images, without relying on movement features. As a result, the performance gained from the depth modality is lower than that of RGB. Nevertheless, the effective fusion of the two modality features can still bring amazing results.
    

    

\begin{table}[!t]
  \centering
    \caption{Comparison of the SOTA methods on the IsoGD.}
  \label{tab:isogd}
  \resizebox{1\linewidth}{!}{
  \begin{tabular}{@{}lccc@{}}
    \toprule
    \textbf{method} & \textbf{modality} & \#\textbf{frames} & \textbf{accuracy(\%)} \\
    \midrule
    c-ConvNet~\cite{wang2018cooperative}  & RGB & 4 &36.60\\
    C3D-gesture~\cite{li2017largeb} & RGB & 32 & 37.28  \\
    AHL~\cite{Hu2018Learning}  &RGB & 32 & 44.88 \\
    3DDSN~\cite{duan2018unified} & RGB & 32 &46.08 \\
    3DCNN+LSTM~\cite{zhang2017learning}  & RGB & 32 & 51.31 \\
    AttentionLSTM~\cite{zhu2019redundancy}  & RGB & 16 & 57.42 \\
    NAS~\cite{yu2021searching} & RGB & 32 & 58.88\\
    Decouple+Recouple~\cite{Zhou_2022_CVPR} & RGB & 64 & 60.87\\
    RAAR3DNet~\cite{Zhou_Li_Wan_2021} & RGB & 64 & 62.66\\
    MSA-3D~\cite{chen2022multi} & RGB & 16 & 62.73\\
    \textbf{UMDR} & RGB & 16/32/64 & 60.60/63.68/\textbf{64.40} \\
    \midrule
    c-ConvNet~\cite{wang2018cooperative}    & Depth & 4 & 40.08\\
    C3D-gesture~\cite{miao2017multimodal}  & Depth & 32 & 40.49  \\
    AHL~\cite{Hu2018Learning} &Depth & 32 & 48.96  \\
    3DCNN+LSTM~\cite{zhang2017learning} & Depth & 32 &49.81 \\
    AttentionLSTM~\cite{zhu2019redundancy}  & Depth & 16 & 54.18 \\
    3DDSN~\cite{duan2018unified} & Depth & 32 &54.95 \\
    NAS~\cite{yu2021searching} & Depth & 32 & 55.68\\
    Decouple+Recouple~\cite{Zhou_2022_CVPR} & Depth & 64 & 60.17\\
    RAAR3D~\cite{Zhou_Li_Wan_2021} &Depth & 64 & 60.66 \\
    MSA-3D~\cite{chen2022multi} & Depth & 16 & 61.72\\
    \textbf{UMDR} & Depth & 16/32/64 & 63.37/64.62/\textbf{65.51} \\
    \midrule
    c-ConvNet~\cite{wang2018cooperative}    & RGB-D & 4 & 44.80\\
    AHL~\cite{Hu2018Learning}   &RGB-D & 32 & 54.14\\
    3DCNN+LSTM~\cite{zhang2017learning} &RGB-D & 32 & 55.29 \\
    AttentionLSTM~\cite{zhu2019redundancy}  & RGB-D & 16 & 61.05 \\
    HybridNet w/o hand~\cite{wang2020hybrid} & RGB-D & 32 & 61.14 \\
    3D-CDCN~\cite{elboushaki2020multid} & RGB-D & 32 & 62.04 \\
    HybridNet w/ hand~\cite{wang2020hybrid} & RGB-D & 32 & 64.61\\
    NAS~\cite{yu2021searching} & RGB-D & 32 & 65.54\\
    RAAR3DNet~\cite{Zhou_Li_Wan_2021} & RGB-D & 64 & 66.62  \\
    Decouple+Recouple~\cite{Zhou_2022_CVPR} & RGB-D & 64 & 66.79\\
    MSA-3D~\cite{chen2022multi} & RGB-D & 16 & 68.15\\
    \textbf{UMDR (CFCer)} & RGB-D & 16/32/64 & 69.23/72.61/\textbf{72.67}  \\
    \bottomrule
  \end{tabular}
  }
\end{table}
\subsection{Comparison with State-of-the-art Methods}\label{sec:sota}
\subsubsection{Results on NTU-RGBD}
Table.\ref{tab:ntu} compares the performance of our method with SOTA methods on the NTU-RGBD
dataset. As the skeleton information is available in this dataset, many recent works tend to perform 2D/3D skeleton-based action recognition on it because the skeleton inherently highlights the key information of the human body, whilst being robust to various illuminations and complex backgrounds. 
However, the generalization ability and robustness of skeleton-based methods are limited. Therefore, we only use color and depth data for action recognition in this paper. 

As can be seen, our method performs the best on both CS and CV protocols (96.2\% on CS and 98.0\% on CV). Meanwhile, comparing with the skeleton-based SOTA method CTR-GCN \cite{chen2021channel}, the proposed method achieves about $\uparrow$3.8\% improvement on CS protocol and $\uparrow$1.2\% on CV protocol. 
And the performance of the single modality even surpasses most skeleton-based approaches \cite{chen2021channel, cheng2020decoupling, 2020Skeleton}, which further demonstrates the robustness of our method to noisy backgrounds and its strong motion perception abilities.
Furthermore, we also compare with other multimodal-based approaches such as Deep Bilinear~\cite{Hu_2018_ECCV}, MMTM~\cite{Joze_2020_CVPR} and PoseC3D~\cite{Duan_2022_CVPR}. These methods use both rgb/rgb-d and skeleton clues for motion recognition, which, undoubtedly, make the network less disturbed by background noise and improve performance. However, our approach achieves competitive performance without resorting to additional fine-grained information, \ie, skeleton.

\begin{table}[!t]
  \centering
    \caption{Comparison of the SOTA methods on the NvGesture. 
    }
  \label{tab:nv}
  \resizebox{1\linewidth}{!}{
  \begin{tabular}{@{}lccc@{}}
    \toprule
    \textbf{method} & \textbf{modality} & \#\textbf{frames} & \textbf{accuracy(\%)} \\
    \midrule
    GPM~\cite{GPM} & RGB & 80 & 75.90\\
    PreRNN~\cite{Yang_2018_CVPR} & RGB & 10 & 76.50\\
    Transformer~\cite{d2020transformer} & RGB & 40 & 76.50\\
    ResNeXt-101~\cite{kopuklu2019real} & RGB & 16/24/32 & 66.40/72.40/78.63\\
    MTUT~\cite{Abavisani_2019_CVPR} & RGB & 64 & 81.33\\
    NAS~\cite{yu2021searching} & RGB & 32 & 83.61\\
    RAAR3DNet~\cite{Zhou_Li_Wan_2021} & RGB & 64 & 85.83\\
    Decouple+Recouple~\cite{Zhou_2022_CVPR} & RGB & 64 & 89.58\\
    \textbf{UMDR} & RGB & 16/32/64 & 83.42/86.25/\textbf{89.80} \\
    \midrule
    Transformer~\cite{d2020transformer} & Depth & 40 & 83.00\\
    ResNeXt-101~\cite{kopuklu2019real} & Depth & 16/24/32 & 72.82/79.25/83.8 \\
    PreRNN~\cite{Yang_2018_CVPR} & Depth & 10 & 84.40\\
    MTUT~\cite{Abavisani_2019_CVPR} & Depth & 64 & 84.85\\
    GPM~\cite{GPM} & Depth & 80 & 85.50\\
    NAS~\cite{yu2021searching} & Depth & 32 & 86.10\\
    RAAR3DNet~\cite{Zhou_Li_Wan_2021} & Depth & 64 & 86.67\\
    Decouple+Recouple~\cite{Zhou_2022_CVPR} & Depth & 64 & 90.62\\ 
    \textbf{UMDR} & Depth & 16/32/64 & 87.25/88.04/\textbf{91.19} \\
    \midrule
    Transformer~\cite{d2020transformer} & RGB-D & 40 & 84.60\\
    PreRNN~\cite{Yang_2018_CVPR} & RGB-D & 10 & 85.00\\
    MTUT~\cite{Abavisani_2019_CVPR} & RGB-D & 64 & 85.48\\
    GPM~\cite{GPM} & RGB-D & 80 & 86.10\\
    MMTM ~\cite{Joze_2020_CVPR} & RGB-D & 32 & 86.31\\
    NAS~\cite{yu2021searching} & RGB-D & 32 & 88.38\\
    \textbf{human} & RGB-D & - & 88.40\\
    RAAR3DNet~\cite{Zhou_Li_Wan_2021} & RGB-D & 64 & 88.59\\
    Decouple+Recouple~\cite{Zhou_2022_CVPR} & RGB-D & 64 & 91.70\\
    \textbf{UMDR (CFCer)} & RGB-D & 16/32/64 & 87.89/90.04/\textbf{92.00}  \\
    \bottomrule
  \end{tabular}
  }
\end{table}
\subsubsection{Results on Chalearn IsoGD}
IsoGD is a much harder dataset because (1) it covers gestures in multiple fields and different motion scales from subtle fingertip movements to large arm swings, and (2) many gestures have high similarity. However, as shown in Table.\ref{tab:isogd}, our approach also achieves the best result on this dataset, \ie, improving about $\uparrow$1.7\% on RGB modality, $\uparrow$3.8\% on depth modality, and $\uparrow$4.5\% on RGB-D modality. It is worth mentioning that the performance of training with 16 frames is about $>$2\% higher than that of the SOTA methods~\cite{Zhou_Li_Wan_2021} and \cite{Zhou_2022_CVPR} trained with 64 frames.

There are two reasons that account for this improvement. On the one hand, the spatiotemporal regularization method can significantly reduce the empirical risk of model on this dataset. On the other hand, decoupling can learn dimension-specific features, while the stage-wise recoupling strategy encourages spatiotemporal compact representation learning. Therefore, the combination of these two advantages can not only strengthen the robustness and generalization but also reinforce the appearance and motion perception of the model.

\begin{table}[!t]
  \centering
    \caption{Comparison of the SOTA methods on the THU-READ.}
  \label{tab:thu}
  \resizebox{1\linewidth}{!}{
  \begin{tabular}{@{}lccc@{}}
    \toprule
    \textbf{method} & \textbf{modality} & \#\textbf{frames} & \textbf{accuracy(\%)} \\
    \midrule
    SlowFast~\cite{feichtenhofer2019slowfast} & RGB & 32 & 69.58\\
    NAS~\cite{yu2021searching} & RGB & 32 & 71.25\\
    TSN~\cite{wang2016temporal} & RGB & 25 & 73.85\\
    Trear~\cite{li2021trear} & RGB & 32 & 80.42\\
    Decouple+Recouple~\cite{Zhou_2022_CVPR} & RGB & 64 & 81.25\\
    \textbf{UMDR} & RGB & 16/32/64 &  81.88/82.50/\textbf{82.94}\\
    \midrule
    TSN~\cite{wang2016temporal} & Depth & 25 & 65.00\\
    SlowFast~\cite{feichtenhofer2019slowfast} & Depth & 32 & 68.75\\
    NAS~\cite{yu2021searching} & Depth & 32 & 69.58\\
    Trear~\cite{li2021trear} & Depth & 32 & 76.04\\
    Decouple+Recouple~\cite{Zhou_2022_CVPR} & Depth & 64 & 77.92\\
    \textbf{UMDR} & Depth & 16/32/64 & 78.42/79.59/\textbf{80.93} \\
    \midrule
    MDNN~\cite{tang2018multi} & RGB+Flow+Depth & - & 62.92\\
    TSN~\cite{wang2016temporal} & RGB+Flow & 25 & 78.23\\
    TSN~\cite{wang2016temporal} & RGB-D+Flow & 25 & 81.67\\
    \midrule
    SlowFast~\cite{feichtenhofer2019slowfast} & RGB-D & 32 & 76.25\\
    NAS~\cite{yu2021searching} & RGB-D & 32 & 78.38\\
    Trear~\cite{li2021trear} & RGB-D & 32 & 84.90\\
    Decouple+Recouple~\cite{Zhou_2022_CVPR} & RGB-D & 64 & 87.04\\
    \textbf{UMDR (CFCer)} & RGB-D & 16/32/64 & 85.80/88.09/\textbf{90.04}  \\
    \bottomrule
  \end{tabular}
  }
\end{table}
\subsubsection{Results on NvGesture}
NvGesture is challenging due to the small size of the data and many fine-grained variations. However, the proposed method performs the best for both single-modal and multimodal evaluating as shown in Table.\ref{tab:nv}, which indicates: (i) the proposed ShuffleMix+ can significantly enlarge the knowledge space of training data, while the proposed UMDR is able to reduce the optimization difficulty of video; (ii) the learned spatiotemporal representations are both discriminative and generalizable, whereby preventing the overfitting problem of models.
Moreover, compared with \cite{Zhou_Li_Wan_2021} and \cite{yu2021searching} that employ NAS to automatically build a gesture-oriented architecture, our approach achieves about $\uparrow$4\% improvement for RGB modality and about $\uparrow$5\% improvement for depth modality, which further demonstrates its generalization ability on gesture recognition.

\subsubsection{Results on THU-READ} 
Table.\ref{tab:thu} compares the performance of our method with SOTA benchmarks on the THU-READ
dataset. Here we demonstrate the effectiveness of the proposed method for scene-based action recognition.
As can be seen, our method exceeds the SOTA results and achieves the best average accuracy (90.04\%) on the released protocol, which further demonstrates that our method is also robust to the complicated background and conversion of view angle.

\section{Conclusion}
In this paper, we improve RGB-D-based motion recognition both from data and algorithm perspectives. First, we propose a novel video data augmentation method based on MixUp that can achieve both spatial and temporal regularization for motion recognition. Second, we introduce a new video modeling method of spatiotemporal decoupling followed by multi-stage recoupling. Finally, we propose to improve multimodal late fusion by exploring modality-specific complementary correspondences over the multimodal semantic information.
We hope our work can inspire more research works on both data- and algorithm-level motion recognition.

\section{Limitations}
Although our method achieves the best performance on four public RGB-D motion datasets, there are still several limitations to UMDR. We summarize and collect them here. Fist, the current version is only capable of isolated motion recognition, meaning that each video corresponds to a single class. However, our method can be extended to continuous motion recognition, where one video can correspond to multiple classes. Because ShuffleMix has the inherent advantage of being able to speculate on class boundaries when processing video sequences mixed along the time dimension, allowing it to identify video clips corresponding to a particular class. This is the work we plan to pursue in the future.
Second, we only explored our method on RGB-D modalities, while other modalities, such as optical flow and infrared, remain to be further validated. Third, due to the relatively heavy computation of the model, the current version may not be suitable for mobile deployment. Therefore, making the model lightweight is the direction of our future efforts.


%



\ifCLASSOPTIONcompsoc
  \section*{Acknowledgments}
\else
  \section*{Acknowledgment}
\fi

This work was supported by the National Key Research and Development Plan under Grant 2021YFE0205700, the External cooperation key project of Chinese Academy Sciences 173211KYSB20200002, the Science and Technology Development Fund of Macau Project 0123/2022/A3, 0070/2020/AMJ, 0004/2020/A1, Guangdong Provincial Key R\&D Programme: 2019B010148001, Open Research Projects of Zhejiang Lab No. 2021KH0AB07, CCF-Zhipu AI Large Model OF 202219, and the Alibaba Group through Alibaba Research Intern Program.

\ifCLASSOPTIONcaptionsoff
  \newpage
\fi


\bibliographystyle{IEEEtran}
\bibliography{IEEEabrv, reference}

%

\begin{IEEEbiography}[{\includegraphics[width=1in,height=1.25in, clip,keepaspectratio]{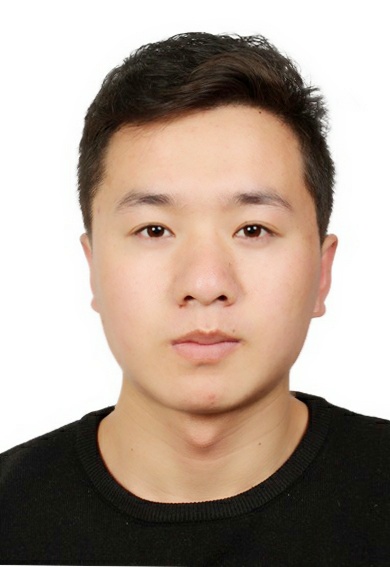}}]{Benjia Zhou} received the B.Eng. degree from the Minzu University of China, Beijing, China, in 2019, and he is currently pursuing a PhD degree in Macau University of Science and Technology, Macau, China.
His research interests focus on action/gesture recognition, sign language recognition/translation, and neural architecture search.
\end{IEEEbiography}
\vspace{-0.2cm}
\begin{IEEEbiography}[{\includegraphics[width=1in,height=1.25in, clip,keepaspectratio]{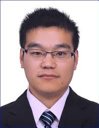}}]{Pichao Wang} received the Ph.D. degree in computer science from the University of Wollongong, Wollongong, NSW, Australia. He is currently a Senior Research Scientist in Amazon Prime Video, USA. He has authored 90+ peer reviewed papers, including those in highly regarded journals and conferences such as IJCV, IEEE TMM, CVPR, ICCV, ECCV, ICLR, AAAI, ACM MM, etc. He is the recipient of CVPR2022 Best Student Paper Award. He is named AI 2000 Most Influential Scholar during 2012-2022 by Miner, due to his contributions in the field of multimedia. He is also in the list of World’s Top 2\% Scientists named by Stanford University. He serves as the Area Chair of ICME 2021, 2022. He also serves as an Associate Editor of Journal of Computer Science and Technology (Tier 1, CCF B).
\end{IEEEbiography}
\vspace{-0.2cm}
\begin{IEEEbiography}[{\includegraphics[width=1in,height=1.25in,clip,keepaspectratio]{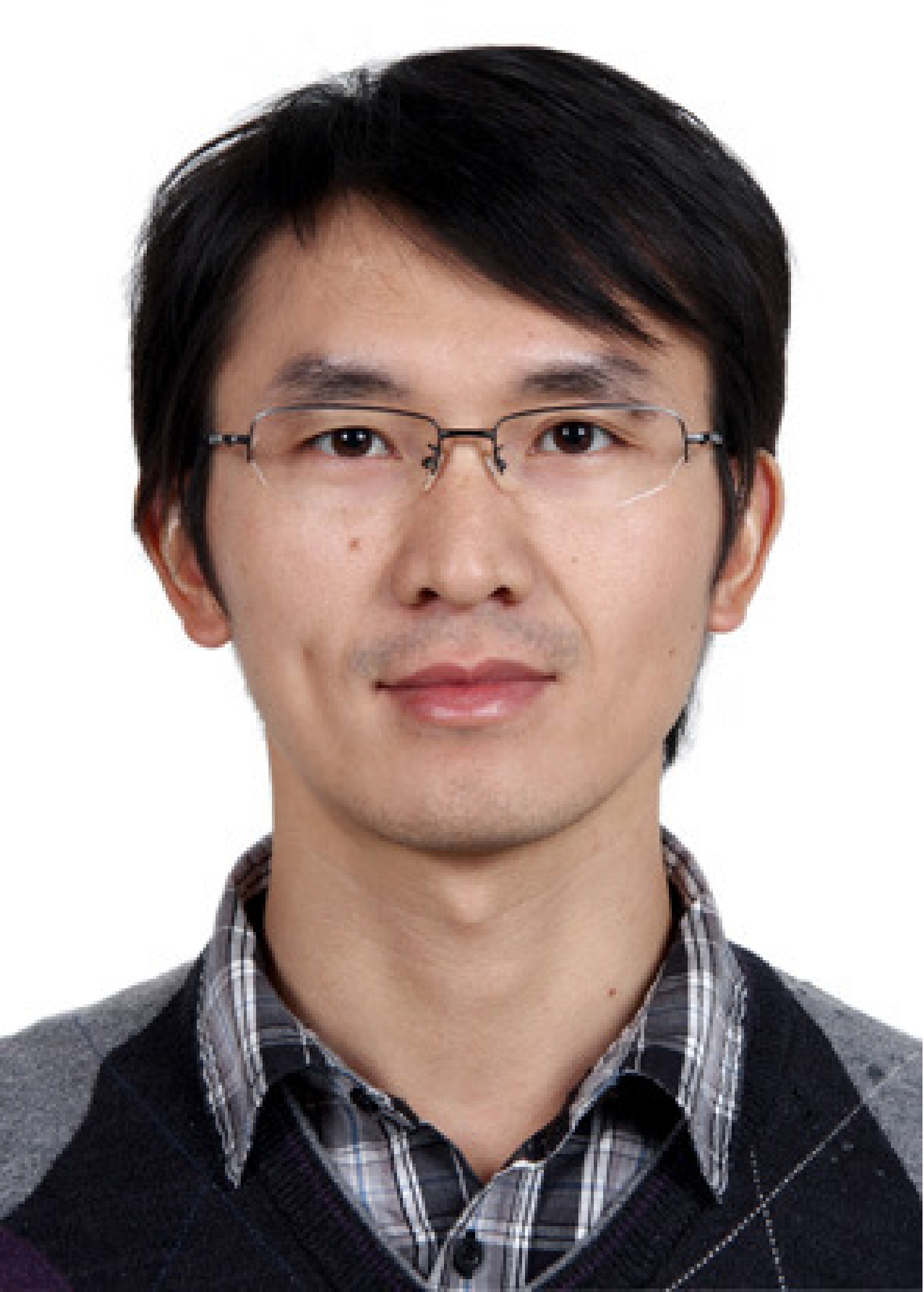}}]{Jun Wan}
received the BS degree from the China University of Geosciences, Beijing, China, in 2008, and the PhD degree from the Institute of Information Science, Beijing Jiaotong University, Beijing, China, in 2015. Since January 2015, he has been a Faculty Member with the National Laboratory of Pattern Recognition, Institute of Automation, Chinese Academy of Science, China, where he currently serves as an Associate Professor. His main research interests include computer vision and machine learning, especially for facial analysis, gesture, and action recognition. He is an associate editor of IET Biometrics. He has served as co-editor of special issues in IEEE TPAMI, IJCV, and IEEE TBIOM.
\end{IEEEbiography}
\vspace{-0.2cm}
\begin{IEEEbiography}[{\includegraphics[width=1in,height=1.25in,clip,keepaspectratio]{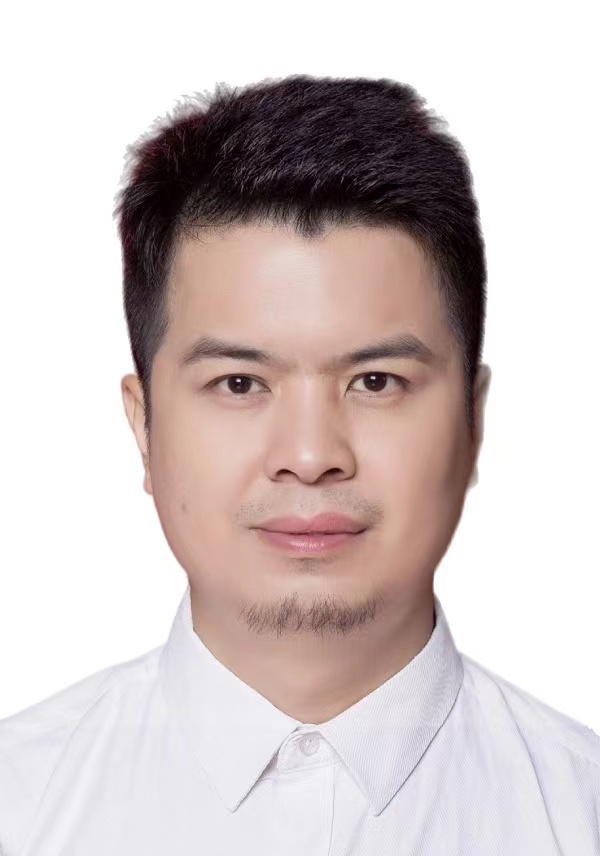}}]{Yanyan Liang} received the B.S. degree from the Chongqing University of Posts and Telecommunications, Chongqing, China, in 2004, and the M.S. and Ph.D. degrees from the Macau University of Science and Technology (MUST), Taipa, Macau, in 2006 and 2009, respectively. He is currently an Associate Professor with MUST. He has published more than 40 papers related to pattern recognition, image processing, and computer vision in IEEE Transactions and international conferences, including TIP, TCYB, TMM, TIFS, TITS, AAAI, IJCAI, CVPR, ICPR, and FG. His research interests include computer vision, image processing, and machine learning.
\end{IEEEbiography}
\vspace{-0.2cm}
\begin{IEEEbiography}[{\includegraphics[width=1in,height=1.25in,clip,keepaspectratio]{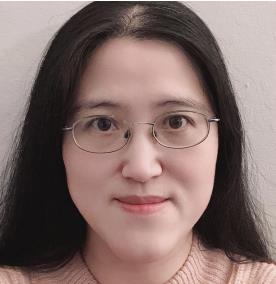}}]{Fan Wang}
received the B.S. and M.S. degree from the Department of Automation, Tsinghua University, Beijing, China, and the Ph.D. degree from the Department of Electrical Engineering, Stanford University, California, United States. She is currently with Alibaba Group as a senior staff algorithm engineer. Her research interests include object tracking and recognition, 3D vision and multi-sensor fusion.
\end{IEEEbiography}






\end{document}